\documentclass[letterpaper, 10 pt, journal, twoside]{IEEEtran}

\usepackage{times}

\usepackage{graphics}           
\usepackage{times}              
\usepackage{amsmath}            
\usepackage{amssymb}            
\usepackage{graphicx}
\usepackage{algorithm}
\usepackage[noend]{algpseudocode}
\usepackage{booktabs}
\usepackage{color}
\usepackage{listings}
\usepackage{subfiles}
\usepackage{hyperref}
\usepackage[nocompress]{cite} 
\definecolor{instructioncolor}{rgb}{.5,.5,.5}

\usepackage[font=small]{caption}

\def\figref#1{Fig.~\ref{#1}}
\def\tabref#1{Tab.~\ref{#1}}
\def\eqref#1{Eq.~(\ref{#1})}

\makeatletter
\usepackage{xspace}
\DeclareRobustCommand\onedot{\futurelet\@let@token\@onedot}
\def\@onedot{\ifx\@let@token.\else.\null\fi\xspace}
 
\def\ie{i.e\onedot}

\makeatother

\usepackage{array}
\newcolumntype{L}[1]{>{\raggedright\let\newline\\\arraybackslash\hspace{0pt}}m{#1}}
\newcolumntype{C}[1]{>{\centering\let\newline\\\arraybackslash\hspace{0pt}}m{#1}}
\newcolumntype{R}[1]{>{\raggedleft\let\newline\\\arraybackslash\hspace{0pt}}m{#1}}

\usepackage{makecell}
\usepackage{pifont}
\usepackage{multirow}
\usepackage[table]{xcolor}
\usepackage{subcaption}
\usepackage{stmaryrd}

\usepackage{flushend}

\begin{document}
\title{\LARGE \bf Globally Consistent RGB-D SLAM with 2D Gaussian Splatting}

\author{Xingguang Zhong$^{1}$, Yue Pan$^{1}$, Liren Jin$^{1}$, Marija Popovi\'{c}$^{2}$, Jens Behley$^{1}$, and Cyrill Stachniss$^{1,3}$}

\twocolumn[{%
\renewcommand\twocolumn[1][]{#1}%

\maketitle

\begin{center}
  \centering
  \vspace{-25pt}
  \captionsetup{type=figure}
  \includegraphics[width=1.0\linewidth]{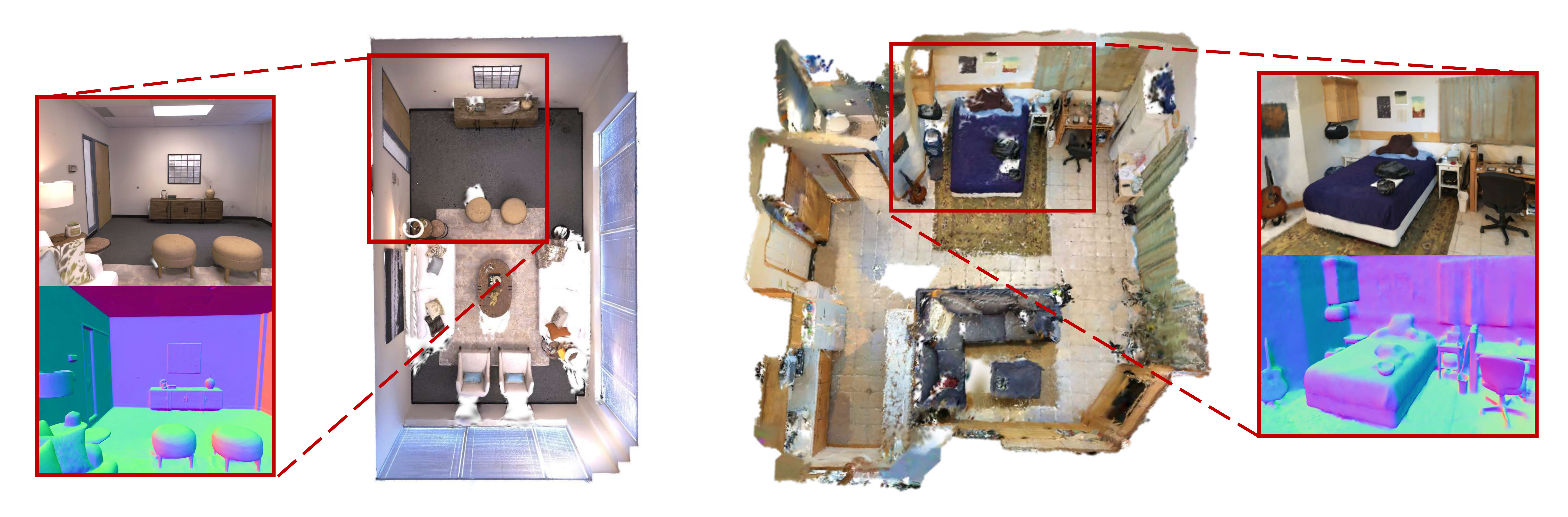}
  \setlength{\abovecaptionskip}{2pt}
  \captionof{figure}{ Reconstruction results of 2DGS-SLAM on synthetic dataset Replica~\cite{straub2019arxiv-Replica} and real-world dataset ScanNet~\cite{dai2017cvpr}.  We present the reconstructed 2D Gaussian splatting maps, along with RGB and normal renderings from zoomed-in local views. These results demonstrate that our method achieves both high-fidelity image rendering and precise geometric reconstruction.} 
  \label{fig:teaser}
  \vspace{0pt}
\end{center}%
}]

\begin{abstract}
  Recently, 3D Gaussian splatting-based RGB-D SLAM displays remarkable performance of high-fidelity 3D reconstruction. 
  However, the lack of depth rendering consistency and efficient loop closure limits the quality of its geometric reconstructions and its ability to perform globally consistent mapping online.
  In this paper, we present 2DGS-SLAM, an RGB-D SLAM system using 2D Gaussian splatting as the map representation. By leveraging the depth-consistent rendering property of the 2D variant, we propose an accurate camera pose optimization method and achieve geometrically accurate 3D reconstruction. 
  In addition, we implement efficient loop detection and camera relocalization by leveraging MASt3R, a 3D foundation model, and achieve efficient map updates by maintaining a local active map.
  Experiments show that our 2DGS-SLAM approach achieves superior tracking accuracy, higher surface reconstruction quality, and more consistent global map reconstruction compared to existing rendering-based SLAM methods,
  while maintaining high-fidelity image rendering and improved computational efficiency.
\end{abstract}

\makeatletter{\renewcommand*{\@makefnmark}{}
\footnotetext{$^{1}$X. Zhong, Y. Pan, L. Jin, J. Behley, C. Stachniss are with the Center for Robotics, University of Bonn, Germany.
$^{2}$M. Popovi\'{c} is with the MAVLab, TU Delft, the Netherlands. 
$^{3}$C. Stachniss is also with the Lamarr Institute for Machine Learning and Artificial Intelligence, Germany.}\makeatother

\vspace{-2pt}
\section{Introduction}
\label{sec:intro}

\IEEEPARstart{S}{imultaneous} localization and mapping (SLAM) is a fundamental problem in computer vision and robotics. 
The ability to reconstruct unknown environments is a basis for various robotic tasks, including navigation~\cite{Maier2012Humanoids, Werby2024rss, Huang2024iser} and exploration~\cite{Bartolomei2023ral, jin2024ral-activegs}.
Recently, radiance field-based map representations like neural radiance field (NeRF)~\cite{mildenhall2020eccv} and Gaussian splatting (GS)~\cite{kerbl2023tog}, have opened up new possibilities for dense RGB-D SLAM by enabling high-fidelity reconstruction with photorealistic rendering. 
Among them, Gaussian splatting has gained popularity due to its fast rendering speed and flexible scalability, establishing itself as the more favorable map representation for radiance field-based RGB-D SLAM.

Most existing GS-based methods~\cite{matsuki2024cvpr-monogs, keetha2024cvpr-splatam, zhu2025threedv-loopsplat, yugay2023arxiv} directly adopt classical 3D Gaussian splatting~(3DGS) for mapping and frame-to-map camera tracking.
However, the depth images rendered from 3DGS at different viewpoints often exhibit inconsistency, negatively impacting pose optimization with depth information and geometric reconstruction accuracy.
Furthermore, since pose drift in long-term camera tracking is inevitable, SLAM systems need to incorporate loop closures as well as map correction and update mechanisms for global consistency~\cite{stachniss2016handbookphoto}.
Some radiance field-based RGB-D SLAM methods~\cite{zhu2025threedv-loopsplat, liso2024cvpr} address this issue by using multiple submaps and applying global transformations to the submaps after loop closure. 
However, they often rely on computationally expensive point cloud registration for relocalization and typically require complex post-processing to merge all submaps, 
making them impractical for online robotic applications.

In this paper, we investigate the problem of realizing a RGB-D SLAM system that builds geometrically accurate and globally consistent radiance field reconstructions online.
Instead of using 3DGS, we adopt 2D Gaussian splatting~(2DGS)~\cite{huang2024siggraph-2dgs} as our map representation. 
2DGS replaces 3D ellipsoids with 2D disks and explicitly computes ray-disk intersections, ensuring consistent depth rendering while maintaining high-fidelity radiance field reconstruction required for novel view synthesis.
Leveraging these properties, we develop an accurate rendering-based method for frame-to-map camera pose estimation. 
In addition, 2DGS represents the environment with discrete Gaussian splats distributed in 3D space, offering a point cloud-like structure allowing for elastic properties when closing loops.
By associating each Gaussian splat with nearby keyframes, we can update the poses of keyframes and their corresponding splats after pose graph optimization. 
Building on this strategy, we further address two key challenges to achieve globally consistent map reconstruction in an online manner.
First, inspired by classical surfel-based dense SLAM methods~\cite{whelan2015rss, behley2018rss}, we maintain a continuously updated local active map and design a mechanism to transition Gaussian primitives between active and inactive states. 
In this way, we prevent tracking and relocalization failures due to the accumulation of new and old map structures, while removing the need of complicated submap management.
Second, after detecting a loop closure, we need to accurately estimate the relative pose between existing frame and the current frame to add proper constraints to the pose graph. 
Unlike prior works that rely on computationally expensive 3D point cloud registration, we leverage MASt3R~\cite{leroy2024eccv}, a recently introduced 3D foundation model with remarkable generalization capability, to estimate an initial relative pose. 
This initial estimate is then refined through further frame-to-map tracking within the active map, achieving accurate relocalization.

The main contribution of this paper is a 2DGS-based RGB-D SLAM system, termed 2DGS-SLAM. Our 2DGS-SLAM addresses two key limitations of existing 3DGS-based SLAM systems. First, to overcome the limited tracking and reconstruction accuracy caused by inconsistent depth rendering in existing 3DGS-based SLAM approaches, we derive a camera pose estimation method specifically adapted to the 2DGS rendering process and implement it efficiently in CUDA. Leveraging the inherent depth rendering consistency of 2DGS, we construct an accurate and robust tracking algorithm and achieve precise surface reconstruction at the same time.
Second, to address the lack of robust loop closure in current systems, we introduce an efficient Gaussian splat management strategy and integrate MASt3R to realize reliable loop closure. This allows for the online reconstruction of globally consistent radiance fields.
As shown in \figref{fig:teaser}, our 2DGS-SLAM achieves outstanding reconstruction results in synthetic dataset and real-world scenes.


In summary, we make three key claims:
(i) Our proposed 2DGS-SLAM achieves superior tracking accuracy compared to state-of-the-art rendering-based approaches;
(ii) Our approach surpasses or is on-par with 3DGS-based methods in surface reconstruction quality and demonstrates more consistent mapping results in real-world scenes compared to other loop-closure-enabled methods. 
At the same time, 2DGS-SLAM maintains high-fidelity image rendering performance that is either superior to or on-par with baseline approaches.
(iii) Compared to other radiance field-based methods that support loop closure, our approach is more efficient on runtime and has a more compact map representation.
These claims are backed up by our experimental evaluation. The open-source implementation of our 2DGS-SLAM is available at:~\url{https://github.com/PRBonn/2DGS-SLAM}.
%

\section{Related Work}
\label{sec:related}

\subsection{Map-centric RGB-D SLAM}
Compared to sparse feature-based visual SLAM systems~\cite{forster2014icra, mur-artal2017tro, campos2021tro}, that target pose and feature location estimation, dense visual SLAM systems generate 3D maps beneficial for robotic tasks like interaction and navigation.
RGB-D SLAM predominates indoor dense SLAM systems, as the depth camera enables direct acquisition of metrically-scaled dense geometry.
Dense visual SLAM systems can be further classified into frame-centric and map-centric approaches based on their tracking strategies. 
Frame-centric methods estimate poses through either sparse feature matching~\cite{endres2014tro, labbe2019jfr} or by minimizing photometric and geometric errors between consecutive frames~\cite{kerl2013icra, kerl2013iros, teed2021neurips, della-corte2018icra}. 
In these methods, the map is merely a by-product constructed by accumulating frame-wise point clouds.
In contrast, similar to LiDAR-based SLAM systems~\cite{kiss2025arxiv, blanco2025ijrr, vizzo2023ral, pan2024tro}, map-centric methods incrementally build a 3D model of the environment and perform frame-to-map tracking for robust pose estimation.

In the past decade, numerous works employ truncated signed distance function (TSDF)~\cite{curless1996siggraph, newcombe2011ismar, whelan2012rssws, dai2017tog, palazzolo2019iros, vizzo2022sensors}, Octomap~\cite{endres2014tro, hornung2013ar}, or surfels~\cite{keller2013threedv, stueckler2014vcir, whelan2015rss, schops2019cvpr} as map representations and use weighted moving average for efficient incremental mapping.
Despite their effective mapping and localization capabilities, these methods suffer from limited scalability and map fidelity, constrained by their discrete map representations.

Recent advancements in radiance fields and implicit neural representations~\cite{park2019cvpr, azinovic2022cvpr, zhong2023icra} have enabled high-fidelity scene modeling, offering new opportunities for map-centric SLAM.
With the radiance field as the map, camera tracking can be performed by minimizating of photometric and geometric discrepancies between the current frame and the rendered image from the radiance field.
%
iMap~\cite{sucar2021iccv} pioneered the use of neural radiance fields (NeRF)~\cite{mildenhall2020eccv} as a map representation, demonstrating the advantages of neural implicit representations in handling the sparse observations or occlusions through inpainting. 
However, despite being memory-efficient, the use of a single multi layer perceptron (MLP) to represent the whole scene limits its ability to capture fine-grained details in complex, large-scale environments.
To improve scalability and rendering performance, subsequent works propose hybrid map representations that combine locally-defined optimizable features with a globally-shared shallow MLP. 
These features can be structured in various forms such as hierarchical voxel grids~\cite{zhu2022cvpr}, octrees~\cite{yang2022ismar}, spatial hashing~\cite{wang2023cvpr-coslam}, tri-plane grids~\cite{johari2023cvpr, deng2024cvpr-plgslam}, or unordered points~\cite{sandstrom2023iccv, liso2024cvpr, zhang2024arxiv-glorie}. 
Nevertheless, the rendering process remains computationally intensive due to ray-wise sampling and volumetric integration.

3D Gaussian splatting~\cite{kerbl2023tog} introduces a novel radiance field based on rasterization of optimizable Gaussian primitives, offering superior training and rendering efficiency while maintaining or exceeding the rendering quality of NeRF.
%
%
These properties have facilitated various robotic applications, such as active sensing~\cite{jin2024ral-activegs}, scene-level mapping~\cite{wei2024ral}, and simulation~\cite{zhou2024cvpr-drivinggs}, thereby encouraging the adoption of 3DGS as the map representation for SLAM.

3DGS-based visual SLAM systems can be split into coupled and decoupled ones, based on whether the online-built 3DGS map is utilized for rendering-based tracking.
Decoupled systems~\cite{huang2024cvpr-photoslam, ha2024eccv, peng2024siggraph,  sandstrom2024arxiv, wu2025arxiv-vingsmono} employ external trackers~\cite{segal2009rss, mur-artal2017tro, campos2021tro, teed2021neurips} for camera pose estimation. 
However, these systems require maintaining a separate map for the external tracker, which is distinct from the 3DGS map, resulting in architectural redundancy in the system design.
In contrast, coupled systems~\cite{matsuki2024cvpr-monogs, yan2024cvpr, keetha2024cvpr-splatam, yugay2023arxiv, sun2024iros-hfslam, Giacomini12025arxiv} utilize 3DGS as the sole map representation for both tracking and mapping through rendering-based gradient descent optimization. 
These systems typically employ a keyframe-based strategy, where mapping is performed using keyframes, while tracking is applied to all frames.
%

%
%
%
%
%

%
Although achieving comparable tracking performance and superior map photorealism to previous map-centric SLAM systems, the aforementioned coupled 3DGS SLAM systems face two main challenges. First, geometric ambiguity in 3D Gaussian splatting limits the accuracy of geometry-based tracking and surface reconstruction. Second, these systems function primarily as visual odometries, lacking the capability to handle loop closures necessary to create a globally consistent map.

To address the first challenge, one solution is to flatten the 3D Gaussian ellipsoids into optimizable 2D surfels, as demonstrated in 2DGS~\cite{huang2024siggraph-2dgs,dai2024siggraph-Gaussian-surfels}. 
2DGS provides enhanced geometric representation with multi-view consistent depth and normal rendering, motivating its use over 3DGS as the map representation to improve geometry-based tracking accuracy and surface reconstruction quality.
While several concurrent works~\cite{wu2025arxiv-vingsmono, huang2025icra-endo2dtam, pan2025rss} adopt 2DGS as their map representation, none have implemented on-manifold camera pose optimization using the 2DGS rasterizer, as MonoGS does for 3DGS.
Our work addresses this gap by explicitly deriving Jacobians for 2DGS-based camera tracking and implementing them in an efficient CUDA-based rasterizer.
In the next section, we discuss related works addressing the second challenge of globally consistent mapping.

\subsection{Visual Loop Closure and Globally Consistent Mapping}

For visual SLAM, closuring loop is crucial for correcting accumulated odometry drift and ensuring a globally consistent map.
Loop closure correction typically involves a place recognition step to identify loop closure candidates, followed by a relocalization step to estimate the relative pose between the current frame and the loop candidate. This relative pose is subsequently used in graph optimization to correct drift errors of trajectory and deform the map.

Compared to distance-based loop candidate search~\cite{kerl2013iros, tang2023tog}, appearance-based place recognition is more versatile, as it can operate without prior knowledge of the camera position and remains effective even when odometry drift is significant.
Early approaches primarily rely on aggregating handcrafted local features using bag-of-words~\cite{galvez2012tro_new, glover2012icra}, random ferns~\cite{glocker2014tvcg}, hamming distance embedding binary search tree~\cite{Giammarino2022iros-mdslam}, or VLAD~\cite{arandjelovic2013cvpr} to build databases for efficient searching and matching~\cite{mur-artal2017tro, campos2021tro,labbe2019jfr, whelan2015rss}, or match image sequences~\cite{milford2012icra, vysotska2016ral}.
Recently, there has been a shift towards learning-based approaches using NetVLAD~\cite{arandjelovic2016cvpr-ncaf} and DINOv2~\cite{oquab2024tmlr, izquierdo2024cvpr-salad, Sergio2024eccv}

The relocalization step aims to estimate the relative pose between the current frame and the detected historical frame.
This transformation serves as a loop constraint edge for pose graph optimization in graph-based SLAM systems.
In cases where odometry drift is small, relocalization becomes a local pose tracking problem, \ie, tracking the current frame against the historical map.
However, for larger loops, where the initial pose often lies outside the convergence basin of pose tracking, a coarse global localization step becomes necessary. 
This is typically achieved using the PnP or Umeyama~\cite{umeyama1991pami} algorithm together with RANSAC, which relies on keypoint-based feature matching~\cite{rusu2009icra, rublee2011iccv}.

Recent data-driven 3D foundation models, particularly DUSt3R~\cite{wang2024cvpr-dust3r} and MASt3R~\cite{leroy2024eccv}, have demonstrated promising performance in various 3D vision tasks. 
Given a pair of RGB images, MASt3R generates a metrically-scaled 3D point map for both images in the first camera's coordinate frame, along with confidence maps. 
From this point map, additional properties including relative camera poses, depth images, and pixel correspondences can be derived.
Furthermore, features extracted by the MASt3R encoder can be aggregated using the ASMK framework~\cite{duisterhof2025threedv} for efficient image retrieval.
Several concurrent SLAM systems leverage MASt3R for different purposes: camera pose and Gaussian splats initialization for 3DGS SLAM~\cite{yu2025icra}, loop closure detection and camera pose tracking~\cite{murai2025cvpr-MASt3Rslam}, and two-view loop constraint construction~\cite{lim2025arxiv}. 
In our approach, we employ MASt3R exclusively for loop closure correction.
Unlike previous approaches that use separate features for loop closure detection and relocalization~\cite{liso2024cvpr, yugay2025cvpr-magicslam}, we utilize MASt3R for both tasks.

Although loop closure correction is a common practice in traditional SLAM systems, it has been adopted by only a few radiance field-based SLAM systems, as it is challenging to maintain a globally consistent radiance field throughout the SLAM process.
%
%
%
Among coupled systems supporting loop closure, most existing approaches utilize a collection of submaps, treating each submap as a rigid body for pose adjustment.
Within each submap, the radiance field can be represented using MLP-based~\cite{tang2023tog}, neural octree-based~\cite{mao2024icra}, or neural point-based~\cite{hu2023neurips, liso2024cvpr} implicit fields, as well as the 3DGS radiance field~\cite{zhu2025threedv-loopsplat, yugay2025cvpr-magicslam}.
While the submap-based strategy is efficient for pose graph optimization and map management~\cite{dai2017tog,reijgwart2019ral,pan2021icra-mvls}, it presents challenges such as drift within the submap and additional effort required for merging submaps and refining the merged map, particularly for the radiance field~\cite{liso2024cvpr, zhu2025threedv-loopsplat, yugay2025cvpr-magicslam}. 
Redundant memory usage occurs in overlapping submap areas, and discrepancies among the submaps are often unfavorable features of the submap-based strategy.

For map representations that are inherently elastic, such as surfels, neural points, and Gaussian splats, one can take a point-based deformation strategy~\cite{whelan2015rss, pan2024tro, pan2025rss} which associates each map primitive with a frame and adjusts frames instead of submaps during pose graph optimization.

As the first coupled 3DGS SLAM system with loop closure, LoopSplat~\cite{zhu2025threedv-loopsplat} employs NetVLAD~\cite{arandjelovic2016cvpr-ncaf} for loop closure detection and estimates loop constraints through rendering-based keyframe-to-submap tracking.
While it achieves superior performance in pose accuracy and map global consistency in larger indoor scenes, the use of 3DGS submaps necessitates a computationally intensive map refinement step after submap merging.
Moreover, without a coarse global localization step, LoopSplat may struggle with relocalization when closing a large loop, where the loop candidate is distant from the current frame.
In contrast, our approach leverages MASt3R for both loop closure detection and coarse relocalization, while adopting a submap-free strategy that associates Gaussian surfels with keyframes.
This design enables direct map correction during camera pose adjustments, avoiding redundant memory usage and submap merging overhead while achieving superior geometric accuracy in a globally consistent 2DGS map.

\begin{figure}[t]
  \centering
  \includegraphics[width=1.0\linewidth]{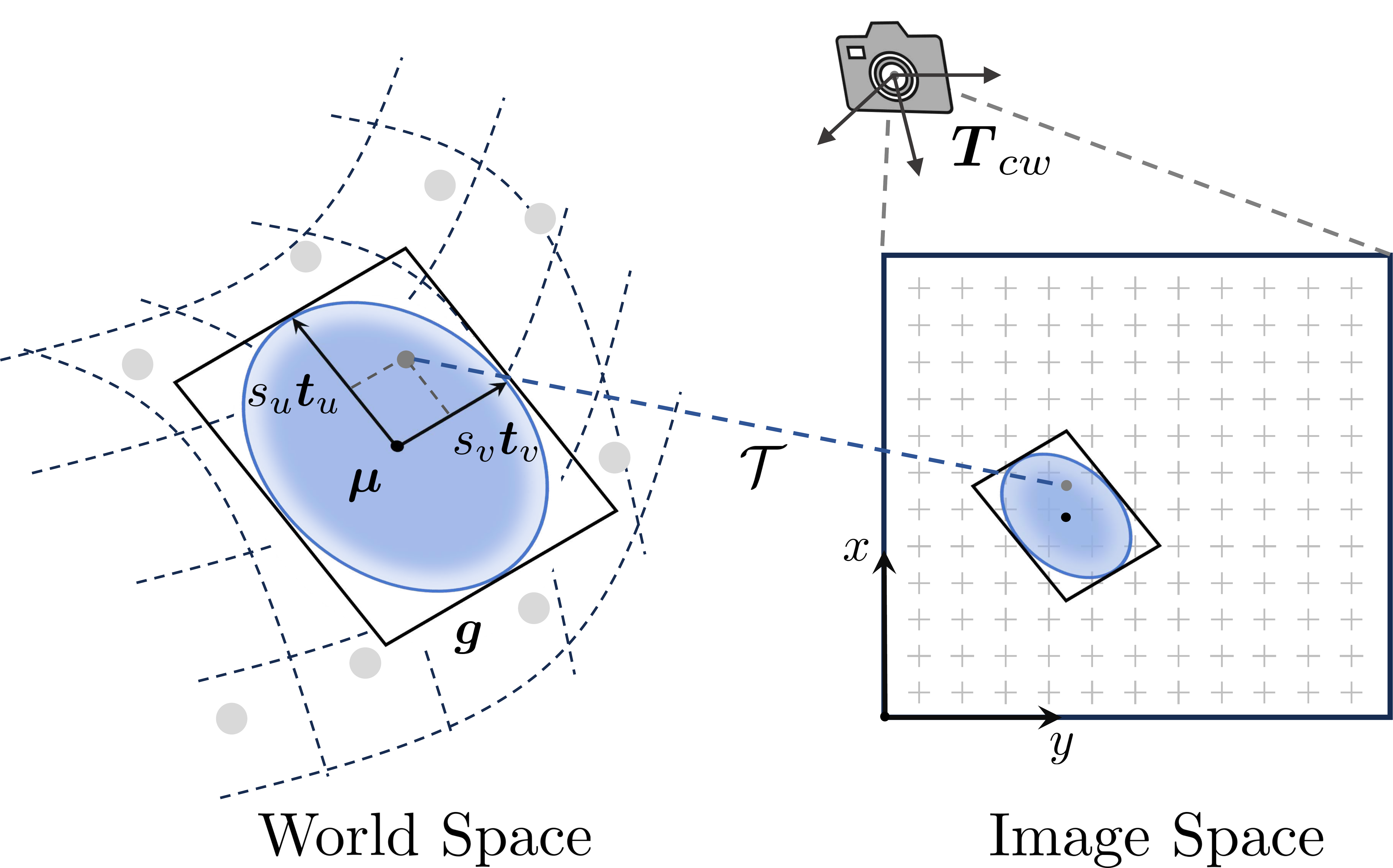}
  \caption{Overview of 2D Gaussian splatting. The pose and shape of a splat $\boldsymbol{g}$ in the world space are defined by its center $\boldsymbol{\mu}$ and
  two scaled tangent vectors $s_u\boldsymbol{t}_u, s_v\boldsymbol{t}_v$. Given a camera with pose $\boldsymbol{T}_{cw}$, the splat $\boldsymbol{g}$ can be projected onto the image space. 
  Points within the local space of the splat is mapped to their corresponding pixel on the image's $x$-$y$ plane via a homography $\mathcal{T}$.}
  \label{fig:2dgs}
\end{figure}


\section{Our Approach}
\label{sec:main}

Our proposed RGB-D SLAM system aims to reconstruct a globally consistent radiance field online while maintaining the precise geometric structure of the 3D environment.  
In the following sections, we first introduce the primary map representation used in our system, 2D Gaussian splatting~\cite{huang2024siggraph-2dgs}, and derive how to backpropagate gradients to the camera pose with 2DGS-based differentiable rendering. 
Next, we describe the structure of our system and provide a detailed explanation of each module in the individual subsections.

\subsection{2D Gaussian Splatting}
\label{sec:2dgs}
Unlike 3DGS, 2DGS compresses one dimension of the 3D ellipsoid to zero, using 2D Gaussian disks as primitives to represent the 3D environment. 
By explicitly calculating the intersection of the rays from the camera with the disk's plane, 2DGS can realize multi-view consistency in depth rendering, thereby achieving a more accurate geometric representation.

As illustrated in \figref{fig:2dgs}, a 2D Gaussian splat $\boldsymbol{g}$ is defined within a local tangent plane in a 3D global coordinate system.
This plane is determined by the splat's central point \(\mbox{\(\boldsymbol{\mu} \in \mathbb{R}^3\)}\) and two principal tangential vectors $\boldsymbol{t}_u$ and $\boldsymbol{t}_v$, with
two scale factors $s_u$ and $s_v$ controlling the variances along the tangential vectors, respectively. 
%
%
%
By representing the rotation matrix of the 2D Gaussian splat as
  $\boldsymbol{R} = \left [ \boldsymbol{t}_u, \boldsymbol{t}_v, \boldsymbol{t}_n\right ] \in \mathbb{R}^{3 \times 3}$,
where \(\mbox{\(\boldsymbol{t}_n = \boldsymbol{t}_u \times \boldsymbol{t}_v\)}\) is the normal vector, 
and arranging scale factors as a $3 \times 3$ diagonal matrix $ \boldsymbol{S} = \mathrm{diag}(s_u, s_v, 0)$,
the 2D local frame can be parameterized as follows:
\begin{equation}
  P(u,v) = \boldsymbol{\mu} + s_u \boldsymbol{t}_u u + s_v \boldsymbol{t}_v v = \boldsymbol{H}\left ( u,v,0,1 \right )^\top,
  \label{eq:Puv}
\end{equation}
\begin{equation}
  \mathrm{where} \; \boldsymbol{H} = \begin{bmatrix}
    s_u\boldsymbol{t}_u& s_v\boldsymbol{t}_v & 0 & \boldsymbol{\mu} \\
    0&  0&  0& 1
  \end{bmatrix} = \begin{bmatrix}
    \boldsymbol{RS} & \boldsymbol{\mu}\\
    0 & 1
   \end{bmatrix}.
\end{equation}

Here, $\boldsymbol{H}$ is the homogeneous transformation matrix from 2D local $uv$ space to the global coordinate system.

The mapping from the $uv$ space to the rendering image's screen space can be formulated as a 2D-to-2D homography transformation~\cite{zwicker2004Graphics}.
Let $\boldsymbol{W} \in \mathbb{R}^{4 \times 4}$ be the transformation matrix from camera space to image space 
and $\boldsymbol{T}_{cw} \in SE(3)$ be the pose of the view camera, combining \eqref{eq:Puv} yields:
\begin{align}
  \boldsymbol{x} = (xz, yz, z, 1)^\top &= \boldsymbol{W} \boldsymbol{T}_{cw} P(u,v) \\
                                    &= \boldsymbol{W} \boldsymbol{T}_{cw} \boldsymbol{H} \left ( u,v,0,1 \right )^\top,
                                    \label{eq:homo}
\end{align}
where $\boldsymbol{T}_{cw}$ transforms the splat in the world space to the camera space and then $\boldsymbol{W}$ transforms it to the image space,
and the $\boldsymbol{x}$ represents the ray corresponding to pixel $(x, y)$ intersecting the 2D Gaussian splat at depth of $z$. 
For convenience, we define:
\begin{align}
  \boldsymbol{H}_c &= \boldsymbol{T}_{cw} \boldsymbol{H} \\
                   &= \boldsymbol{T}_{cw} \begin{bmatrix} \boldsymbol{RS} & \boldsymbol{\mu}\\
                                            0 & 1
                                          \end{bmatrix} = \begin{bmatrix}
                                            s_u\boldsymbol{t}_{uc}& s_v\boldsymbol{t}_{vc} & 0 & \boldsymbol{\mu}_c \\
                                            0&  0&  0& 1
                                          \end{bmatrix},
  \label{eq:hc}
\end{align}
where $\boldsymbol{t}_{uc}$, $\boldsymbol{t}_{vc}$ and $\boldsymbol{\mu}_c$ are Gaussian splat's tangential vectors and central point in camera space.
Furthermore, we define the whole homography $\mathcal{T}$ as:
\begin{equation}
  \mathcal{T} = \boldsymbol{W} \boldsymbol{T}_{cw} \boldsymbol{H} = \boldsymbol{W} \boldsymbol{H}_c\,.
  \label{eq:wth}
\end{equation}

To render the value of pixel $\boldsymbol{p} =(x, y)^\top$ from the splats, 2DGS solves the inverse problem of \eqref{eq:homo}, 
computing the intersection of ray $\boldsymbol{x}$ with the 2D Gaussian splat in $uv$ space, while avoiding the need to compute the inverse of $\mathcal{T}$.
For further details, we refer the reader to the original paper~\cite{huang2024siggraph-2dgs}.

Apart from these geometric parameters mentioned above, each splat also contains color feature $\boldsymbol{c}$ and opacity $\alpha$ to represent its visual appearance.
After computing the ray-splat intersections of all splats within field of view, 2DGS sorts them by depth and uses volumetric alpha blending to integrate weighted appearance values $\boldsymbol{V}_{\boldsymbol{p}}$ of pixel $\boldsymbol{p}$, as follows:
\begin{equation}
  \boldsymbol{V}_{\boldsymbol{p}} = \sum_{i = 0}^{N} \boldsymbol{v}_i \alpha_i \mathcal{G}(\boldsymbol{u}_i^{\boldsymbol{p}}) \prod_{j=0}^{i-1}\left( 1-\alpha_j \mathcal{G}(\boldsymbol{u}_j^{\boldsymbol{p}}) \right)
  \label{eq:alpha_blending},
\end{equation}
where $ \mathcal{G}(\boldsymbol{u})= \mathcal{G}(u, v) = \exp\left ( - \frac{u^2+v^2}{2}  \right ) $ represent the Gaussian weight of the intersection $\boldsymbol{u}$ in the $uv$ space,
$\boldsymbol{u}_i^{\boldsymbol{p}}$ means the $i$-th intersection along the ray of pixel $\boldsymbol{p}$, and $N$ denotes the number of Gaussian splats that intersect with the ray. 
It should be noted that the appearance value $\boldsymbol{v}$ can be a view-dependent color generated from color feature $\boldsymbol{c}$, depth~$d$ and normal vector $\boldsymbol{t}_{nc} = \boldsymbol{t}_{uc} \times \boldsymbol{t}_{vc}$, 
meaning that the color image, depth image and the normal image are rendered in the same way. In addition, we can also render an opacity image $\boldsymbol{O}$ if we set $\boldsymbol{v} = 1$.

To summarize, each 2D Gaussian splat $\boldsymbol{g}$ contains parameters $\left( \boldsymbol{\mu}, \boldsymbol{R}, \boldsymbol{t}_u, \boldsymbol{t}_v, s_u, s_v, \boldsymbol{c}, \alpha \right)$ to describe its geometric and visual information.
These parameters can be progressively optimized through differentiable rasterization using a rendering loss to achieve high-fidelity reconstruction. 
2DGS implements both the forward rendering and backward gradient propagation in CUDA, enabling efficient and scalable operation.

\subsection{Camera Pose Optimization}
\label{sec:cameraopt}

Our proposed SLAM system does not rely on external visual odometry. Instead, we directly use rendering-based frame-to-map tracking to estimate the pose of each frame.
The core problem of rendering-based tracking is computing the gradient of the rendering loss with respect to the camera pose.
However, similar to 3DGS, the original 2DGS assumes that the camera poses of input frames are fixed and the loss generated from the forward rendering cannot propagate to them.
Some 3DGS-based SLAM systems~\cite{yugay2023arxiv, keetha2024cvpr-splatam, zhu2025threedv-loopsplat} apply the pose matrix directly to all Gaussian splats, 
and derive the gradient with respect to each element of the matrix by automatic differentiation.
They then leverage differentiable transformation between quaternion and rotation matrix to obtain the quaternion's gradient.
However, these methods 
cannot guarantee that the gradient remains in $SE(3)$ during the optimization process, 
resulting in a method that is neither efficient nor accurate.

To address this limitation, MonoGS~\cite{matsuki2024cvpr-monogs} derives analytical Jacobians of camera pose in $SE(3)$ for 3DGS and achieves efficient tracking.
However, due to the difference of rendering mechanism, this derivation cannot be transfered to 2DGS directly. In our work, we bridge the gap and derive the camera Jacobians explicitly based on Lie algebra for 2DGS.
To save memory overhead, the 2DGS map in our system does not use a spherical harmonic function to generate view dependent colors, so spherical harmonic function is not considered in the derivation below.

Since both the ray-splat intersection and alpha blending-based rendering are differentiable, given the loss $L$ generate between rendered image and input image, 
the per-element gradients of $L$ with respect to the homography $\mathcal{T}$, denoted as $\frac{\partial L}{\partial \mathcal{T}}$, can be obtained from 2DGS's original implementation.
Based on \eqref{eq:hc} and \eqref{eq:wth}, we can derive the gradients of $\boldsymbol{H}_c$ from $\frac{\partial L}{\partial \mathcal{T}}$ by applying the chain rule:
\begin{align}
  \frac{\partial L}{\partial \boldsymbol{H}_c} =
  \begin{bmatrix}
    \frac{\displaystyle \partial L}{\displaystyle s_u \partial \boldsymbol{t}_{uc}} & 
    \frac{\displaystyle \partial L}{\displaystyle s_v \partial \boldsymbol{t}_{vc}} & 
    0 & 
    \frac{\displaystyle \partial L}{\displaystyle \partial \boldsymbol{\mu}_c}\\
    0 & 0 & 0 & 0
  \end{bmatrix} =
  \boldsymbol{W}^\top \frac{\partial L}{\partial \mathcal{T}}
  \,.
\end{align}

From this gradient matrix, we can directly extract the gradients with respect to $\boldsymbol{t}_{uc}$, $\boldsymbol{t}_{vc}$, and $\boldsymbol{\mu}_c$, which are given by $\frac{\partial L}{\partial \boldsymbol{t}_{uc}}$, $\frac{\partial L}{\partial \boldsymbol{t}_{vc}}$, and $\frac{\partial L}{\partial \boldsymbol{\mu}_{c}}$.
Furthermore, according to \eqref{eq:alpha_blending}, 2DGS can render normal images from splats' normal vector $\boldsymbol{t}_{nc}$ in the camera space. 
Then, the gradient of loss $L$ with respect to $\boldsymbol{t}_{nc}$, i.e., $\frac{\partial L}{\partial \boldsymbol{t}_{nc}}$, can be computed through the backpropagation of alpha blending. 
Combining these results, we obtain the full gradient of $\boldsymbol{R}_c$:
\begin{equation}
  \frac{\partial L}{\partial \boldsymbol{R}_c} = \left [ \frac{\partial L}{\partial \boldsymbol{t}_{uc}}, \frac{\partial L}{\partial \boldsymbol{t}_{vc}}, \frac{\partial L}{\partial \boldsymbol{t}_{nc}} \right ]\,.
\end{equation}

The camera pose $\boldsymbol{T}_{cw}$ affects the rendered image by transforming each 2D Gaussian splat from world space to camera space.
This transformation impacts both the center $\boldsymbol{\mu}$ and the orientation $\boldsymbol{R}$ of the splat, producing their transformed counterparts $\boldsymbol{\mu}_c$ and $\boldsymbol{R}_c$ in the camera coordinate system.
Accordingly, the gradient of $\boldsymbol{T}_{cw}$ is composed of two distinct components:
\begin{equation}
  \frac{\partial L}{\partial \boldsymbol{T}_{cw}} = \frac{\partial L}{\partial \boldsymbol{\mu}_{c}} \frac{\mathcal{D}\boldsymbol{\mu}_{c} }{\mathcal{D} \boldsymbol{T}_{cw}} \oplus
                                                        \frac{\partial L}{\partial \boldsymbol{R}_{c}} \frac{\mathcal{D}\boldsymbol{R}_{c} }{\mathcal{D} \boldsymbol{T}_{cw}},
\label{eq:derivatives}
\end{equation}
where $\oplus$ ensures that both terms are projected into the same tangent space of $SE(3)$ before summation, guaranteeing dimensional consistency. Adopting the same notation as in MonoGS~\cite{matsuki2024cvpr-monogs}, we define the partial derivative on the manifold as:
\begin{equation}
\frac{\mathcal{D}f(\boldsymbol{T}) }{\mathcal{D} \boldsymbol{T}} = \lim_{\tau  \to 0}\frac{\mathrm{Log}(f(\mathrm{Exp}(\tau)\circ \boldsymbol{T})\circ f(\boldsymbol{T})^{-1})}{\tau},
\end{equation}
where $\boldsymbol{T} \in SE(3)$ and $\tau \in \boldsymbol{se}(3)$, $\circ$ is a group composition operation. Then the two derivatives in \eqref{eq:derivatives} can be derived as following:
\begin{equation}
  \frac{\mathcal{D} \boldsymbol{\mu}_c }{\mathcal{D} \boldsymbol{T}_{cw}} = \begin{bmatrix}
    I& -\boldsymbol{\mu}_c^\times
  \end{bmatrix},
  \frac{\mathcal{D} \boldsymbol{R}_c }{\mathcal{D} \boldsymbol{T}_{cw}} = \begin{bmatrix}
    0 & -\boldsymbol{R}_{c, :,1}^\times  \\
    0 & -\boldsymbol{R}_{c, :,2}^\times \\
    0 & -\boldsymbol{R}_{c, :,3}^\times
  \end{bmatrix},
\end{equation}
where $\times$ denotes the skew symmetric matrix of a 3D vector, and ${:,i}$ refers to the $i$-th column of the matrix. 
To ensure computational efficiency, we implement the above process in CUDA as well.

\subsection{System Overview}

\begin{figure*}[htbp]
  \centering
  \includegraphics[width=1.0\linewidth]{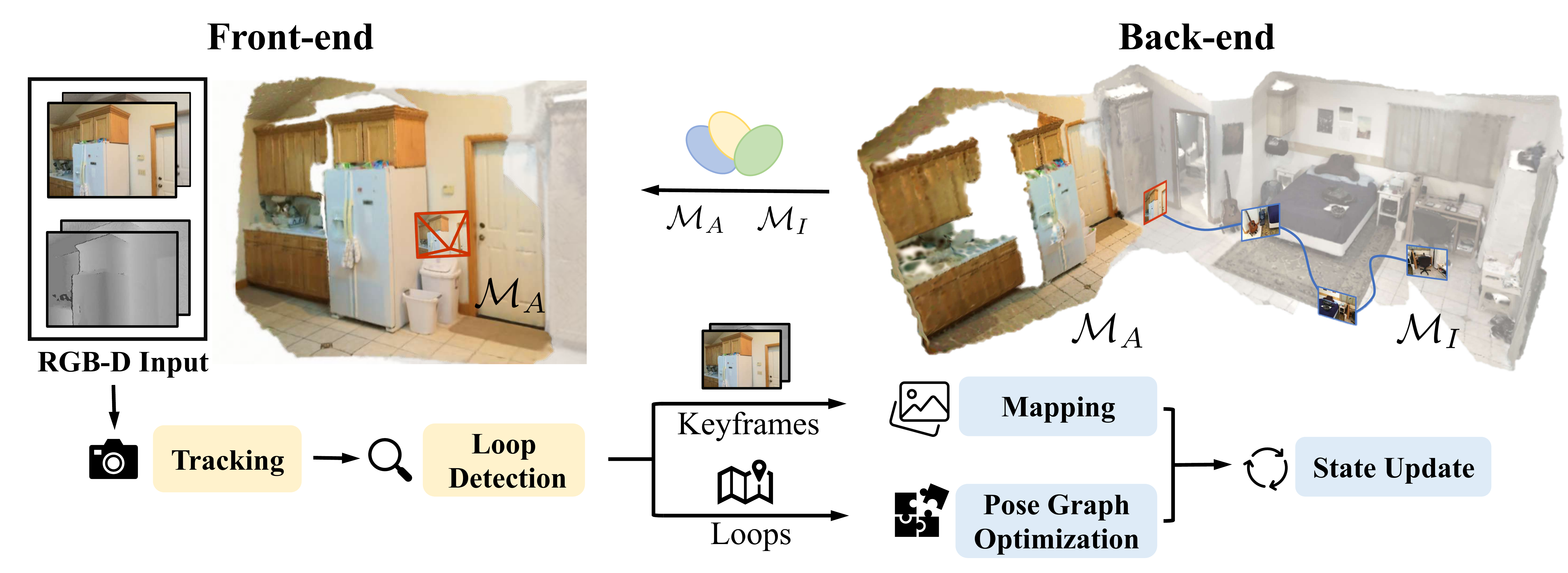}
  \caption{System overview of 2DGS-SLAM. Our system consists of two parallel processes: a front-end and a back-end.
  Taking RGB-D frames as input, the front-end performs frame-to-map camera tracking using the currently active map $\mathcal{M}_A$, and searches for potential loop closures.
  The selected keyframe is sent to the back-end, which uses it to expand and optimize the map (mapping). If a loop closure is detected in the front-end, we send the computed loop constraint to the back-end, where pose graph optimization and map correction will be performed.
  After mapping or pose graph optimization, the back-end updates the active state of the map and synchronizes it with the front-end.
  When no message is received, the back-end keeps refining the map based on previously stored keyframes.}
  \label{fig:overview}
\end{figure*}

\label{sec:overview}
Leveraging the depth-consistent rendering capability of 2DGS, we develop a RGB-D SLAM system to enable accurate camera pose estimation alongside geometrically precise radiance field reconstruction.
To further achieve online reconstruction of a globally consistent map, we extend the parameterization of 2D Gaussian splats. More specifically, our map representation can be expressed as:
\begin{equation}
  \mathcal{M} = \left \{ \boldsymbol{g}_i, \delta_i, t_i^c, d_i^c, t_i^l \mid i=1,...,N \right \},
  \label{eq:map}
\end{equation}
where $\boldsymbol{g}_i = \left( \boldsymbol{\mu}, \boldsymbol{R}, \boldsymbol{t}_u, \boldsymbol{t}_v, s_u, s_v, \boldsymbol{c}, \alpha \right)$ is the original Gaussian splat's learnable parameters as discussed in Sec.~\ref{sec:2dgs}.\\
The item $t_i^c$ denotes the sequential ID of the frame that observed $\boldsymbol{g}_i$ at the closest distance, which is used to associate~$\boldsymbol{g}_i$ with  its corresponding keyframe $t_i^c$ for global map correction, and the closest distance is stored as $d_i^c$.
Meanwhile, $t_i^l$ denotes the ID of the last frame that observed $\boldsymbol{g}_i$. 
The Boolean variable $\delta_i = \{0,1\}$ represents $\boldsymbol{g}_i$'s active state. Based on this state, we can split all the Gaussian splats in the map $\mathcal{M}$ into two subsets $\mathcal{M}_{A} = \{ \delta_i = 1 \mid \boldsymbol{g}_i \in \mathcal{M}\}$ and $\mathcal{M}_I = \{ \delta_i = 0 \mid \boldsymbol{g}_i \in \mathcal{M}\}$, representing active and inactive Gaussian splats respectively.

As shown in \figref{fig:overview}, our system comprises two main process: the front-end and the back-end.
The front-end is responsible for estimating the current camera pose and detecting potential loop closures. 
The back-end focuses on expanding and optimizing the map using frames with estimated poses, updating the active map, as well as globally deforming the map after a loop closure.
We summarize the main components of our system as follows:
\begin{enumerate}
\item Tracking and keyframe selection (Sec.~\ref{sec:Tracking}): In the front-end, upon acquiring a new RGB-D frame, we estimate the camera pose using the currently active map $\mathcal{M}_A$ through a frame-to-map tracking approach. 
Then, we select keyframes based on covisibility and send them to the back-end for mapping.
\item Mapping (Sec.~\ref{sec:Mapping}): We expand the map by projecting 2D Gaussian splats into the world space based on keyframes. The active map $\mathcal{M}_A$ is then optimized for a few iterations using both current and historical keyframes. Afterwards, we send both $\mathcal{M}_A$ and $\mathcal{M}_I$ to the front-end for synchronization. Additionally, $\mathcal{M}_A$ and $\mathcal{M}_I$ are continuously refined using a selection of historical keyframes in the back-end, respectively.

\item Map state update (Sec.~\ref{sec:state_update}): To prevent outdated map regions from negatively impacting tracking, we mark Gaussian splat $\boldsymbol{g}_i$ that have not been observed for a certain period as inactive, i.e., $\delta_i = 0$. 
After a loop closure, we reactivate the observed inactive Gaussian splats to avoid redundancy.

\item Loop detection and relocalization (Sec.~\ref{sec:Loop detection}): In the front-end, we identify loop closures by comparing each incoming frame with previous keyframes. If a candidate keyframe is found, we estimate the relative pose between this keyframe and current frames based on MASt3R~\cite{leroy2024eccv}. 
The pose is then sent to the back-end and used to add a loop constraint to the pose graph. 

\item Map correction (Sec.~\ref{sec:correction}): After receiving loop constraint from the front-end, we perform pose graph optimization, and transform every Gaussian according to the updated pose of its associating keyframe $t_i^c$. Finally, we reactivate observed inactive Gaussian splats and send the deformed map to the front-end.

\end{enumerate}

The following sections will provide more comprehensive explanations of each component.

\subsection{Tracking and Keyframe Selection}

\label{sec:Tracking}
Based on the derivation in \ref{sec:cameraopt}, we can optimize the current camera's pose $\boldsymbol{T}_{cw}$ through gradient descent once the rendering loss is known.
Given the input RGB image $\boldsymbol{I}$ and depth image $\boldsymbol{D}$, we define the color rendering loss $\boldsymbol{L}_{I} \in \mathbb{R}^{H \times W}$ and depth rendering loss $\boldsymbol{L}_{D} \in \mathbb{R}^{H \times W}$ as:
\begin{equation}
  \boldsymbol{L}_{I} = \left \| \boldsymbol{I}_r - \boldsymbol{I}\right \|_{1},
\end{equation}
\begin{equation}
  \boldsymbol{L}_{D} = \left \| \boldsymbol{D}_r - \boldsymbol{D}\right \|_{1},
  \label{eq:LD}
\end{equation}
where $\boldsymbol{I}_r$ and $\boldsymbol{D}_r$ are RGB image and depth image rendered from active submap $\mathcal{M}_A$ at pose $\boldsymbol{T}_{cw}$,
and $\left \| \cdot \right \|_{1}$ means element-wise L1 distance. 

According to \eqref{eq:alpha_blending}, 2DGS can render normal vector images $\boldsymbol{N}_r \in \mathbb{R}^{H \times W \times 3}$ from 2D Gaussian splats.
Utilizing this property, we can filter out the influence of back-facing splats on tracking by applying a normal mask $\boldsymbol{M}_n$, which can be calculated by:
\begin{equation}
  \boldsymbol{M}_n(x,y) = \llbracket \boldsymbol{N}_r(x,y)^\top  \boldsymbol{r}(x,y) > 0 \rrbracket ,
  \label{eq:normal_mask}
\end{equation}
where $\boldsymbol{r}(x,y) \in \mathbb{R}^3$ represents the normalized ray vector emitted from the camera's optic center and passing through pixel $(x,y)$ on the image plane,
and $\llbracket \cdot \rrbracket$ is the indicator function returning 1 if the statement is true, otherwise 0.
Note that all vectors in \eqref{eq:normal_mask} are defined in the camera space. 
Besides, we apply another mask  $\boldsymbol{M}_o$ obtained from the rendered opacity image $\boldsymbol{o} \in \mathbb{R}^{H \times W} $ to ignore the loss generated from under-reconstructed area.
The opacity mask is defined as:
\begin{equation}
  \boldsymbol{M}_o(x,y) = \llbracket \boldsymbol{O}(x,y) > 0.95 \rrbracket.
  \label{eq:opacity_mask}
\end{equation}

Then, the total tracking loss $L_{t} \in \mathbb{R}$ can be written as:

\begin{equation}
  L_{t} = \frac{1}{|\Omega|} \sum_{\boldsymbol{p} \in \Omega} \boldsymbol{M}_n(\boldsymbol{p}) \cdot \boldsymbol{M}_o(\boldsymbol{p}) \cdot (\boldsymbol{L}_{I}(\boldsymbol{p}) + \lambda_d \boldsymbol{L}_{D}(\boldsymbol{p})),
  \label{eq:tracking_loss}
\end{equation}
where $\Omega = \left \{ (u,v) \mid u \in {1,...,W}, v \in {1,...,H}\right \}$ represent all the pixels and $\lambda_d$ is the weight used to balance these two losses, and $\cdot$ represents the per-element product. 
We directly initialize the optimization using last frame's pose and employ the AdamW~\cite{loshchilov2018iclr} algorithm to iteratively optimize the pose until convergence or reaching the maximum number of iterations $n_{\text{iter}}$. 
To ensure that the gradients are stable, during the optimization process of camera tracking, the parameters of these Gaussian splats remain fixed.
It should be noted that, during tracking, we only render images from the active Gaussian splats, i.e., $\boldsymbol{g} \in \mathcal{M}_A$.

As with most SLAM frameworks, rather than using all frames for mapping, we selectively choose keyframes to improve efficiency. 
Similar to MonoGS~\cite{matsuki2024cvpr-monogs}, we primarily determine keyframes based on the covisibility. 

First, from \eqref{eq:alpha_blending}, the alpha-blending coefficient of Gaussian splat $\boldsymbol{g}_k$ in the rendering of pixel $\boldsymbol{p}$ is:
\begin{equation}
  w_{k}(\boldsymbol{p}) = \alpha_k \mathcal{G}(\boldsymbol{u}_k^{\boldsymbol{p}}) \prod_{j=0}^{k-1}(1-\alpha_j \mathcal{G}(\boldsymbol{u}_j^{\boldsymbol{p}})).
  \label{eq:weight}
\end{equation}

We define the contribution of $\boldsymbol{g}_k$ to a rendered frame from a given view $V$ as the sum of its rendering coefficients across all pixels, expressed by:
\begin{equation}
  \mathcal{C}_{k}^{V} = \sum_{\boldsymbol{p}\in \Omega}w_k(\boldsymbol{p})
  \label{eq:contribution}\,,
\end{equation}
where $\boldsymbol{p}\in \Omega$ denotes all pixels.
Intuitively, $\mathcal{C}_{k}^{V}$ represents how many pixels the Gaussian splat $\boldsymbol{g}_k$ contributes to the rendering.
Therefore, we directly define that $\boldsymbol{g}_k$ is visible in the given view $V$  if its $\mathcal{C}_{k}^{V}$ is larger than 0.5.
Furthermore, Given two camera views $A, B$ and current active map $\mathcal{M}_A$, we define the covisibility score between these two views as follows:

\begin{equation}
  S_{cov}(A,B) = \frac{\left | G_A\cap G_B \right | }{\left |G_A\cup G_B \right |},
\end{equation}
where \( G_A = \{\boldsymbol{g}_k \in \mathcal{M}_A \mid \mathcal{C}_k^{A} > 0.5\} \) and similarly for \( G_B \) with \(\mathcal{C}_k^{B}\). They are the sets of all visible Gaussian splats at view $A$ and $B$, respectively.
If the covisibility score between the current view and the last keyframe falls below a threshold $c_k$, 
or if their distance between the translation vectors of $\boldsymbol{T}_{cw}$ surpasses a threshold $d_k$, the current view is selected as a keyframe. 

\subsection{Mapping}
\label{sec:Mapping}

After completing pose estimation in the front-end, the new keyframe $K_n$ observed by the robot is sent to the back-end process for map expansion and optimization.
To reduce memory consumption, we first convert the RGB-D data into a colored point cloud $\mathcal{P}$, and then apply random downsampling to obtain $\mathcal{P}_{s}$ before projecting it into the world space.
Each 3D point is then initialized as a Gaussian splat, where its initial scales $s_u$ and $s_v$ are determined by the distance $d$ to its nearest neighbor in $\mathcal{P}_{s}$, and the opacity $\alpha$ is set to an initial value of 0.99.
The initial normal vector $\boldsymbol{t}_n$ is obtained from the normal image $\boldsymbol{N}_{\boldsymbol{D}}$, which is computed from the pixel gradient of the depth image~$\boldsymbol{D}$.
Specifically, we derive $\boldsymbol{N}_{\boldsymbol{D}}$ from the cross product of neighboring pixel differences in~$\boldsymbol{D}$,
\begin{equation}
  \boldsymbol{N}_{\boldsymbol{D}}(x,y) = \frac{\nabla_x\boldsymbol{D}(x,y) \times \nabla_y\boldsymbol{D}(x,y)}{\left | \nabla_x\boldsymbol{D}(x,y) \times \nabla_y\boldsymbol{D}(x,y) \right | }
  \label{eq:normal_fromD}.
\end{equation}

Here, we assign $\boldsymbol{t}_n$ to each splat based on its corresponding position. We than randomly initialize two principal tangential vectors $\boldsymbol{t}_u$ and $\boldsymbol{t}_v$, which are perpendicular to $\boldsymbol{t}_n$.
In addition,  the closest frame ID $t_i^c$ and last observed frame ID $t_i^l$ of the new splat are both initialized as the ID of the current keyframe.

To prevent the generation of excessively redundant Gaussian splats, we maintain a voxel hash table with resolution~$r_{h}$ for the active submap $\mathcal{M}_A$, which represents the spatial occupancy state. 
New Gaussian splats are only created in spatially unoccupied voxels where no existing Gaussians are present.
This voxel hash table is updated whenever the map expands by adding new splats and when modifications occur in the active state, such as transitioning Gaussians between active and inactive states.
Due to its relatively low resolution and restriction to active regions, the memory overhead remains minimal.

In the back-end process, we continuously optimize all the Gaussian splats $\{ \boldsymbol{g} \in \mathcal{M} \}$ to ensure they not only produce high-fidelity image renderings but also align well with the actual surface, 
accurately capturing the geometric structure of the environment. To achieve this, we train the map with multiple loss functions. Firstly, the color image rendering loss $\boldsymbol{L}_{c} \in \mathbb{R}^{H \times W}$ is expressed as:
\begin{equation}
  \boldsymbol{L}_{c} = \lambda_c \left \| \boldsymbol{I}_r - \boldsymbol{I}\right \|_{1} + (1-\lambda_c) L_{\boldsymbol{SSIM}}(\boldsymbol{I}_r, \boldsymbol{I}),
\end{equation}
where $\lambda_c \in [0,1]$ and the $L_{SSIM}$ represents the structural similarity index measure (SSIM) \cite{wang2004tip}.
We also apply an L1 loss $\boldsymbol{L}_{D}$ like \eqref{eq:LD} to directly supervise the depth rendering optimization using the input depth image.
Following 2DGS, to ensure that the Gaussian splats conform to the surface locally, we add a normal consistency loss $\boldsymbol{L}_{n} \in \mathbb{R}^{H \times W}$ between the rendered depth image $\boldsymbol{D}_r$ and the rendered normal image $\boldsymbol{N}_r$, formulated as:
\begin{equation}
  \boldsymbol{L}_{n} =  \boldsymbol{1}_{H \times W} - \boldsymbol{N}_{\boldsymbol{D}_r} \cdot \boldsymbol{N}_r,
\end{equation}
where $\boldsymbol{N}_{\boldsymbol{D}_r} \in \mathbb{R}^{H \times W \times 3}$ denotes the normal image estimated from the rendered depth image $\boldsymbol{D}_r$ by applying \eqref{eq:normal_fromD}, and \(\cdot\) indicates a per-pixel 3D vector dot product.
%
%
%
%

Finally, we optimize the map $\mathcal{M}$ using the combination of above loss functions, which can be written as:
\begin{equation}
  L_{m} = \frac{1}{|\Omega|} \sum_{\boldsymbol{p} \in \Omega}( \boldsymbol{L}_{c}(p) + w_{d} \boldsymbol{L}_{D}(p) + w_{n} \boldsymbol{L}_{n}(p)), 
\end{equation}
where $w_{d}, w_{n}$ are weights to balance the contributions of the corresponding loss terms.
Utilizing the loss function $L_{m}$, we continually optimize the active map $\mathcal{M}_A$ and inactive map~$\mathcal{M}_I$ separately in the back-end. For $\mathcal{M}_A$, we maintain an active frames set~$\mathcal{S}_a$, which is defined as:
\begin{equation}
  \mathcal{S}_a =  \{t_i^c  \mid  \boldsymbol{g_i} \in \mathcal{M}_A\},
\end{equation}
where $t_i^c$, as described in \eqref{eq:map}, is the ID of the closest observing frame of $\boldsymbol{g}_i$. In each iteration, we randomly sample~$N_a$ frames from $\mathcal{S}_a$ and $N_i$ frames from the other frames to perform optimization for Gaussian splats in $\mathcal{M}_A$ and $\mathcal{M}_I$, respectively.

\subsection{Map State Update}

\label{sec:state_update}

Due to the accumulation of tracking errors, directly integrating newly observed data into the global map can cause a misalignment between new and existing structures, negatively impacting re-localization after loop detection.
To mitigate this issue, as mentioned in Sec.~\ref{sec:overview}, we maintain two separate maps: an active map $\mathcal{M}_A$, which stores recently observed Gaussian splats, and an inactive map $\mathcal{M}_I$, which preserves historical splats. 

Given the latest posed keyframe $K_n$ sent from the front-end, where $n$ is its sequential ID, we first assign active status $(\delta = 1)$ to all new Gaussian splats generated from~ $K_n$ and add them to active map~$\mathcal{M}_A$.
Using the visibility criterion defined in~\eqref{eq:contribution}, we identify which Gaussian splats in~$\mathcal{M}_A$ are visible from the view of~$K_n$ and then update their last observed frame ID to $t_{i}^{l} = n$.
Meanwhile, we compute the distance from these Gaussian splat to the viewpoint of~$K_n$. If the distance is smaller than the historical minimum distance~$d_i^c$, we update their closest frame ID as $t_i^c = n$.
For all $\boldsymbol{g}_i \in \mathcal{M}_A$, we mark $\boldsymbol{g}_i$ as inactive $(\delta = 0)$ if $(n - t_i^l)$ exceeds a predefined time threshold, indicating that $\boldsymbol{g}_i$ has not been observed for a long time.

As illustrated in \figref{fig:active_state}, to avoid accumulating redundant Gaussian splats in the same region, we reactivate inactive splats in $\mathcal{M}_I$ when the robot revisits previously observed areas. 
More specifically, if a loop closure is detected between the current frame $\boldsymbol{F}_c$ and a historical keyframe $\boldsymbol{K}_h$, we first perform pose graph optimization followed by map correction. For the subsequent $T$ keyframes, where $T$ is a predefined hyperparameters, if an inactive Gaussian splat $\boldsymbol{g}_i \in \mathcal{M}_I$ is observed, and its closed observed frame ID $t_i^c > h$, indicating that its position has been corrected by pose graph optimization, we reassign its state as active and update its last observed frame ID $t_i^l$ to the current frame ID $c$.
In addition, during the mapping process, we continuously sample historical keyframes, such as $\boldsymbol{K}_r$, and evaluate the contributions of the Gaussian splats associated with $\boldsymbol{K}_r$, i.e., $\boldsymbol{g} \in \{ t_i^c = r \mid \boldsymbol{g}_i \in \mathcal{M}\}$, based on ~\eqref{eq:contribution}.
If the contribution of a Gaussian splat falls below 0.5, we consider it occluded by surrounding splats and remove it from the map to maintain map compactness.

\begin{figure}[!t]
  \centering
  \begin{subfigure}[b]{0.238\textwidth}
    \centering
    \includegraphics[width=\textwidth]{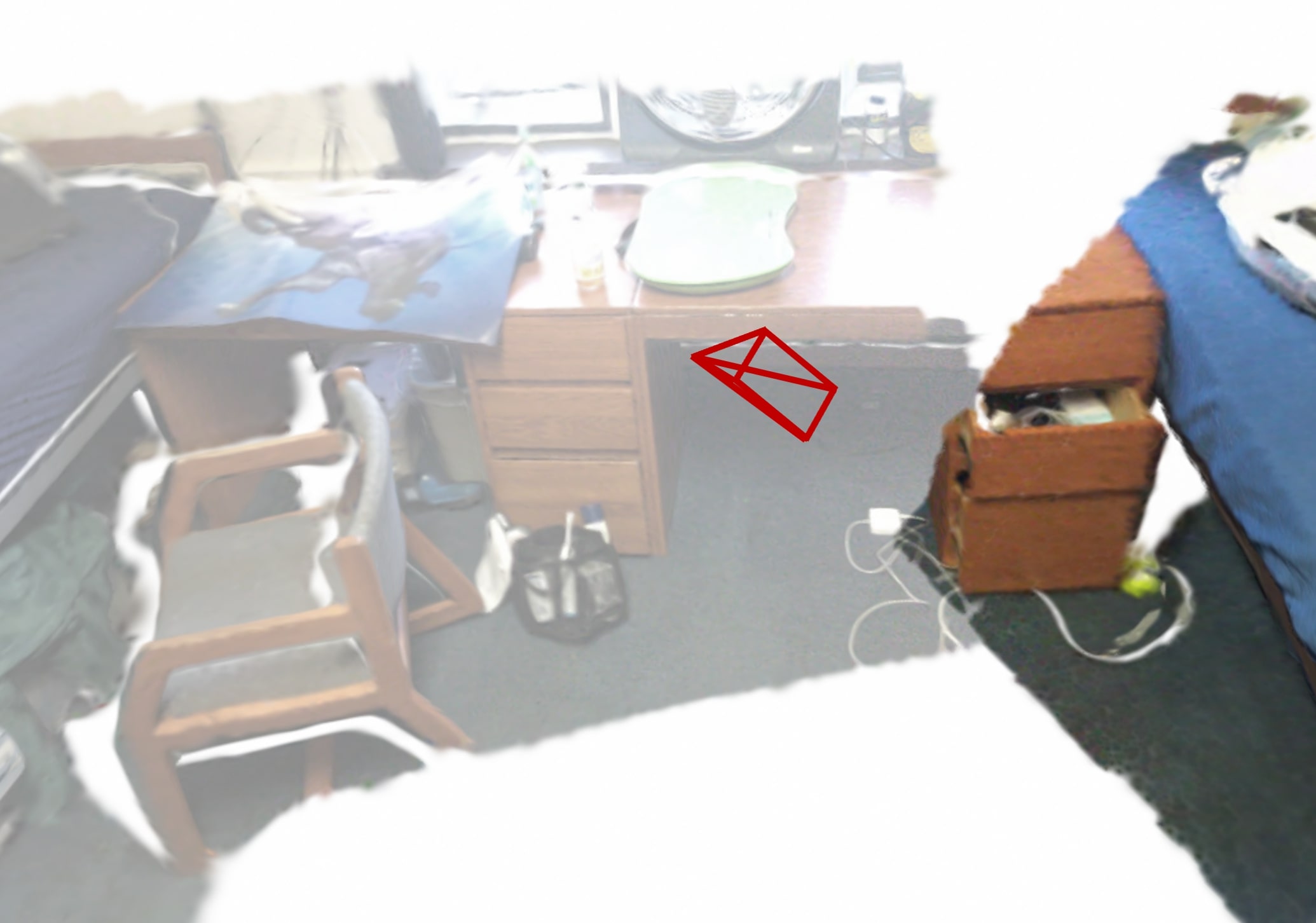}
    \caption{}\vspace{-1pt}
    \label{subfig:a}
  \end{subfigure}
  \hfill
  \begin{subfigure}[b]{0.238\textwidth}
    \centering
    \includegraphics[width=\textwidth]{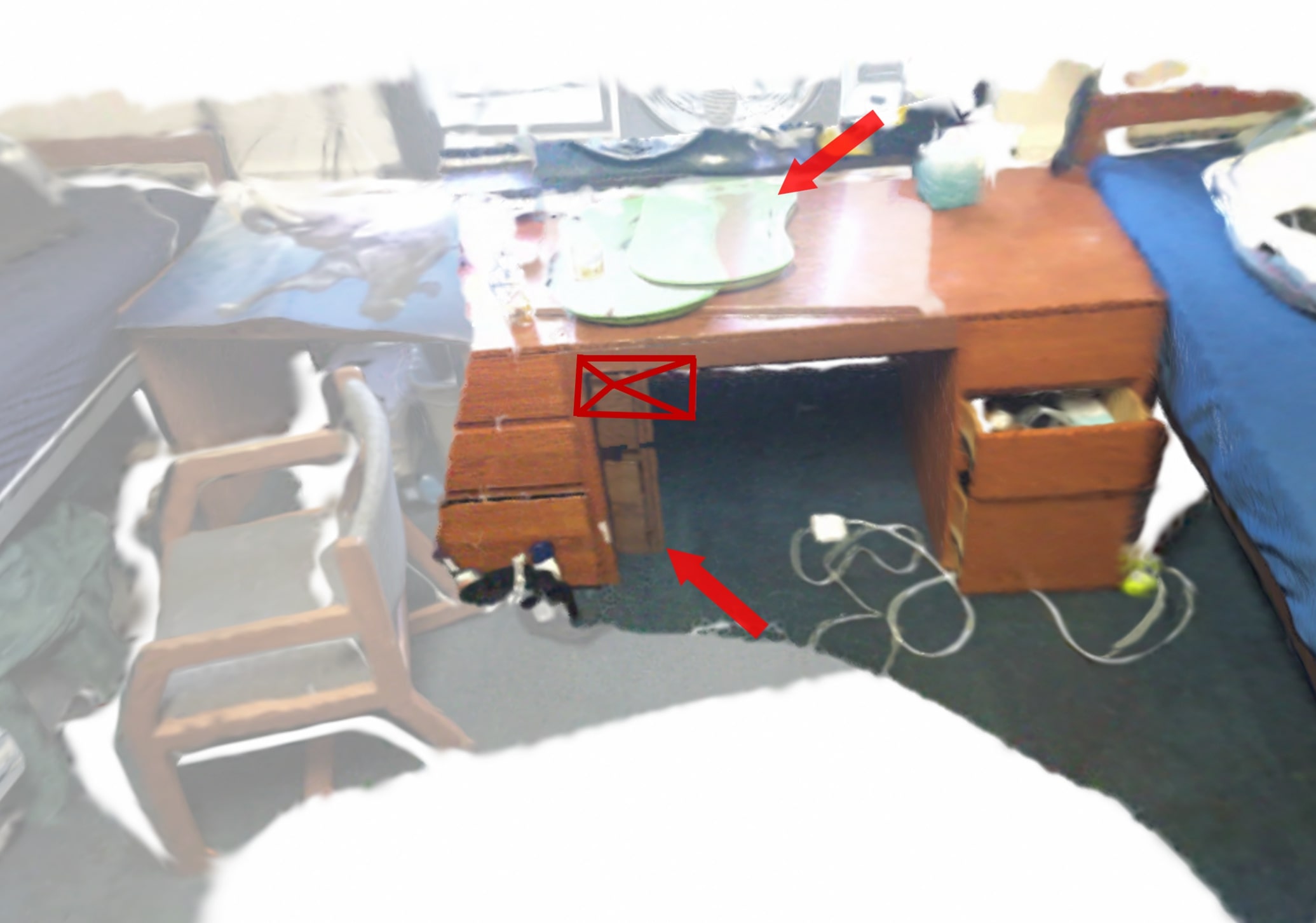}
    \caption{}\vspace{-1pt}
    \label{subfig:b}
  \end{subfigure}


  \begin{subfigure}[b]{0.238\textwidth}
    \centering
    \includegraphics[width=\textwidth]{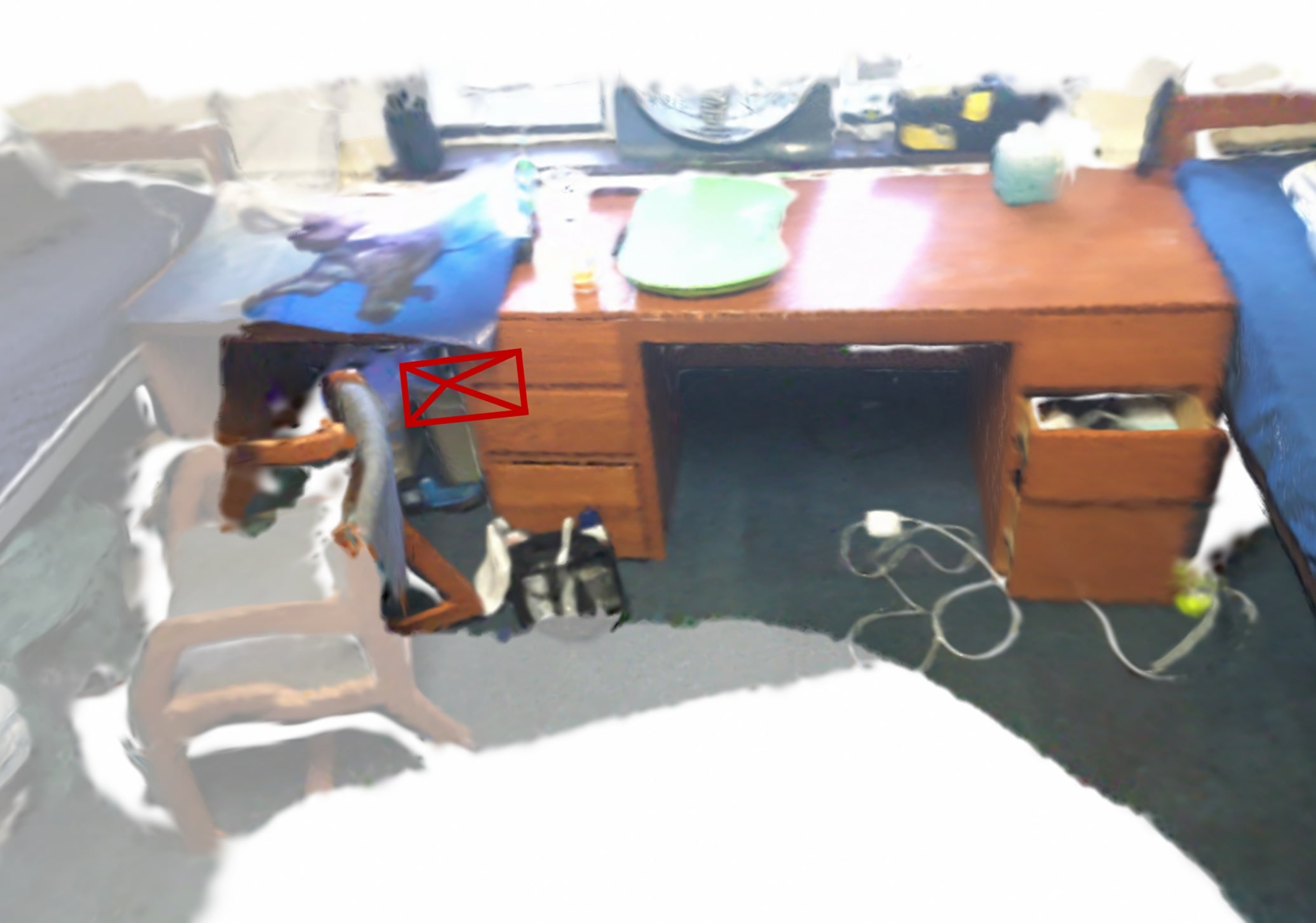}
    \caption{}\vspace{-1pt}
    \label{subfig:c}
  \end{subfigure}
  \hfill
  \begin{subfigure}[b]{0.238\textwidth}
    \centering
    \includegraphics[width=\textwidth]{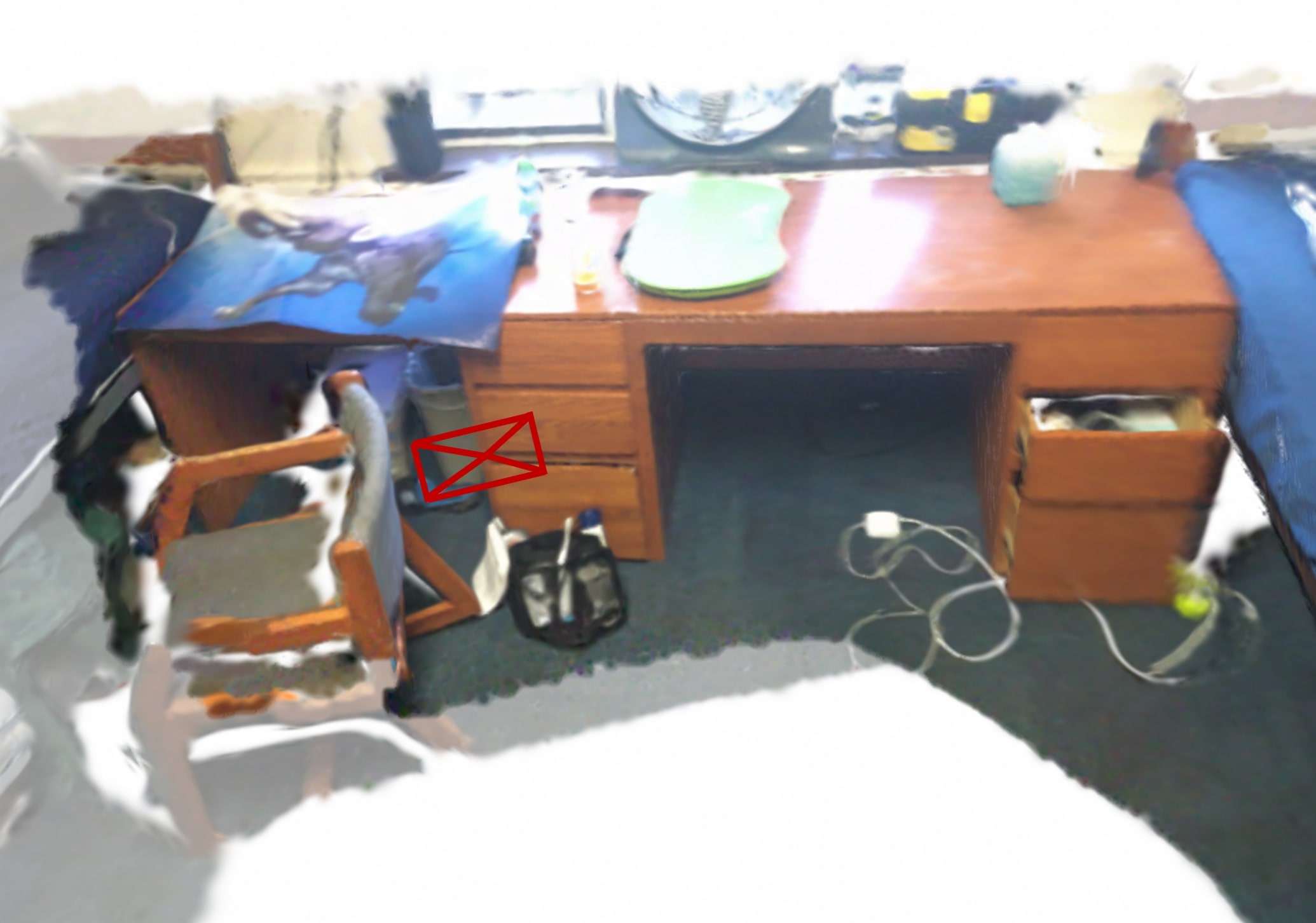}
    \caption{}\vspace{-1pt}
    \label{subfig:d}
  \end{subfigure}

  \caption{Illustration of the state update process. Images (a)-(d) are shown in temporal order. The bright regions represent the active map, while the dim regions indicate the inactive map. The red cones represent the current camera views.
  (a) Active and inactive maps during running. (b) The camera observes part of the inactive map, but no loop closure has been detected. Due to pose drift, the active and inactive maps misalign (see red arrows).
  (c) The system detects a loop closure, aligns inactive and active maps, and reactivates the observed inactive Gaussian splats. 
  (d) As the camera moves, more inactive Gaussian splats are progressively reactivated.}
  \vspace{-10pt}
  \label{fig:active_state} 
\end{figure}

\subsection{Loop Detection and Relocalization}
\label{sec:Loop detection}

In the back-end, we maintain a pose graph $\boldsymbol{G}$, where each keyframe serves as a vertex, and the relative pose between adjacent keyframes forms an edge in the graph.
When a loop closure is detected, we compute the relative pose between the current frame and the candidate frame searched from all the keyframes to introduce a loop closure constraint for pose graph optimization.

We primarily utilize MASt3R~\cite{leroy2024eccv} for loop detection and re-localization. Given a pair of input RGB images $\left \langle \boldsymbol{I}_i, \boldsymbol{I}_j \right \rangle$, 
MASt3R extracts their image features $\mathcal{F}_i, \mathcal{F}_j$ through a vision transformer-based model~\cite{dosovitskiy2021iclr} and directly outputs pixel-wise point clouds, $\boldsymbol{P}_i$ and $\boldsymbol{P}_j$, along with their respective confidence maps, $\boldsymbol{C}_i$ and $\boldsymbol{C}_j$.
Notably, both $\boldsymbol{P}_i$ and $\boldsymbol{P}_j$ are represented in the camera coordinate frame of view $i$. By leveraging these dense point clouds, we can estimate the camera parameters of the two frames and subsequently solve for their relative pose using the PnP algorithm~\cite{Lepetit2009ijcv},
thereby obtaining the depth maps $\boldsymbol{D}_i^p$ and $\boldsymbol{D}_j^p$ in their respective camera space.

Inspired by recent works \cite{duisterhof2025threedv, murai2025cvpr-MASt3Rslam}, we use features $\mathcal{F}$ from the vision transformer encoder as local descriptors, and employ the aggregated selective match kernel (ASMK)~\cite{tolias2013iccv} for image retrieval. 
ASMK quantizes and binarizes these features using a precomputed k-means codebook, producing high-dimensional sparse binary representations.
The similarity between images can be efficiently computed via a kernel function over shared codebook elements. We integrate this process into our online system. 
For each keyframe $K_n$, we utilize MASt3R’s feature encoder to extract image feature $\mathcal{F}_n$ and store it, along with the corresponding image sequence ID $n$, in a feature database managed using ASMK. 

\begin{figure}[t]
  \centering
  \includegraphics[width=\linewidth]{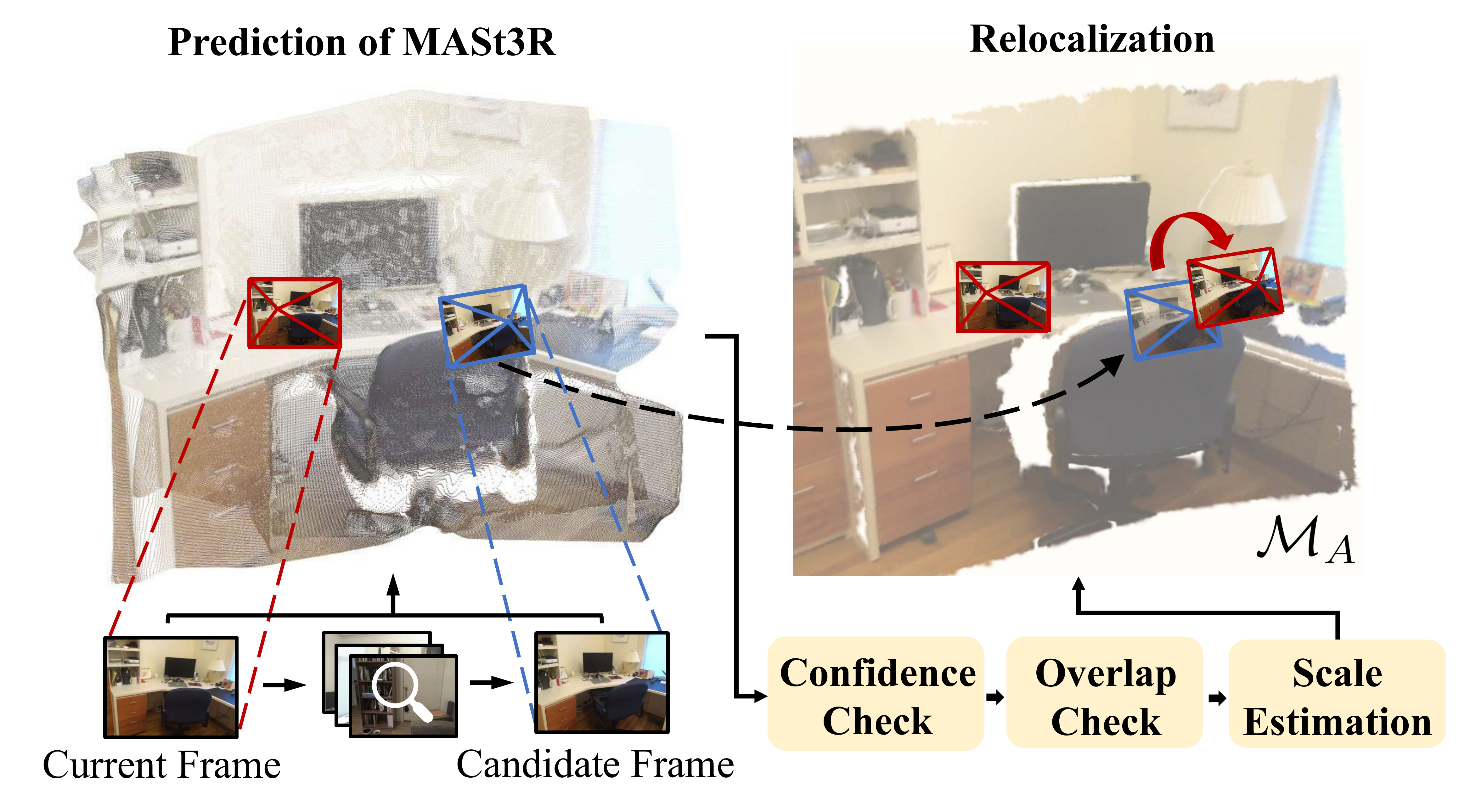}
  \caption{We input the current frame and the loop candidate into MAsT3R to estimate their relative pose and dense point clouds. After confidence and overlap checks, we optimize the scale of computed relative pose by aligning the point cloud to the local map. The scaled pose is then transfered to the world space and refined by tracking it on the active map $\mathcal{M}_A$.}
  \label{fig:loop_closure}
\end{figure}

\figref{fig:loop_closure} illustrates the main pipeline of our loop closure detection.
After tracking the current image $\boldsymbol{I}_c$, we extract its features and compute similarities with all keyframes. 
We then identify the keyframe with the highest similarity to~$\boldsymbol{I}_c$. If its similarity score exceeds a predefined threshold $s_r$, we designate it as a loop closure candidate.
Let $\boldsymbol{I}_l$ and~$\boldsymbol{D}_l$ denote the RGB and depth images of the candidate keyframe~$\boldsymbol{K}_l$, respectively. To further validate the loop closure, 
we feed the image pair $\left \langle \boldsymbol{I}_c, \boldsymbol{I}_l \right \rangle$ to MASt3R, obtaining the predicted point clouds $\boldsymbol{P}_c$~and~$\boldsymbol{P}_l$, the corresponding depth maps~$\boldsymbol{D}_c^p$~and~ $\boldsymbol{D}_l^p$, 
their confidence maps~$\boldsymbol{C}_c$ and~$\boldsymbol{C}_l$, and the estimated relative pose~$\boldsymbol{T}_{lc}$.
If the mean of confidence map $\boldsymbol{C}_c$ is below a predefined threshold $c_s$, we consider the prediction unreliable and discard this loop closure candidate.
Otherwise, we estimate the overlap between the two frames as follows. Since the predicted point clouds are both expressed in the coordinate frame of $\boldsymbol{I}_c$, 
we directly reproject~$\boldsymbol{P}_l$ onto the image plane of $\boldsymbol{I}_c$ and compare the reprojected depth $\boldsymbol{D}_c^l$ with the predicted depth $\boldsymbol{D}_c^p$.
Inspired by~\cite{chen2020rss}, the overlap ratio $O_{lc}$ is computed as:
\begin{equation}
  O_{lc} = \frac{\sum_{\boldsymbol{u} \in \mathcal{V}} \mathbf{1} \left( \left| \boldsymbol{D}_c^{l}(\boldsymbol{u}) - \boldsymbol{D}_c^{p}(\boldsymbol{u}) \right| < \tau_d \right)}{|\mathcal{V}|},
\end{equation}
where $\mathcal{V}$ denotes the pixels falling within the image boundaries after projection, $\mathbf{1}$ is an indicator function that returns~1 if the condition inside is true and 0 otherwise. $\tau_d$ is a depth consistency threshold, which is 0.05 in our setting. 
If $O_{lc}$ is smaller than a predefined threshold $\delta_o$, we determine that the candidate frame lacks sufficient overlap for reliable relocalization and reject the loop closure attempt.

If the candidate passes the filtering, we further compute the accurate relative pose to provide a loop closure constraint for pose graph optimization.
Although MASt3R is trained on a large amount of metric-scale data, its predicted depth maps remain up to scale. 
Therefore, we first estimate the scale factor $s^*$ using real observed depth image $\boldsymbol{D}_c$, given by:
\begin{equation}
   s^* = \arg\min_{s} \left \| \boldsymbol{C}_c \cdot (\boldsymbol{D}_c - s \boldsymbol{D}_c^p) \right \|_2 \,,
\end{equation}
where $\cdot$ represents the per-element product for matrices. This is a weighted least squares problem that can be solved in closed form. Then, we multiply this scale factor with the translation component of $\boldsymbol{T}_{lc}$ to obtain the scaled relative pose $\boldsymbol{T}_{lc}^r$.
Consequently, we derive the candidate keyframe's pose in the world space as $\boldsymbol{T}_l = \boldsymbol{T}_{lc}^r \boldsymbol{T}_c$, where $\boldsymbol{T}_c$ is the pose of the current camera.
To obtain a more accurate estimation, we use $\boldsymbol{T}_l$ as the initial estimate and perform a scan-to-model tracking in current active map $\mathcal{M}_A$. 
After convergence or reaching the maximum number of iterations $n_\mathit{iter}$, we re-render a depth image $\boldsymbol{D}_r$ from the active map at the optimized pose $\boldsymbol{T}_l^{t}$ and compute its L1 error $e_{t}$ against the input depth image $\boldsymbol{D}$
using the same formulation as \eqref{eq:tracking_loss}:
\begin{equation}
  e_{t} = \frac{1}{|\Omega|} \sum_{\boldsymbol{p} \in \Omega} \boldsymbol{M}_n(\boldsymbol{p}) \cdot \boldsymbol{M}_o(\boldsymbol{p}) \cdot \left \| \boldsymbol{D}_r(\boldsymbol{p}) - \boldsymbol{D}(\boldsymbol{p})\right \|_{1},
  \label{eq:tracking_error}
\end{equation}
where $\boldsymbol{M}_n$ and $\boldsymbol{M}_o$ are normal and opacity masks, respectively. Only frame with an error $e_{t}$ below a threshold~$\varepsilon_t$ retained for further refinement.
Then, the successfully optimized pose $\boldsymbol{T}_l^{t}$ is used to caculate the accurate relative pose as $\boldsymbol{T}_{lc}^{t} = \boldsymbol{T}_l^{t} \boldsymbol{T}_c^{-1}$. 
With $\boldsymbol{T}_{lc}^{t}$ as the loop constraint, we perform pose graph optimization and update the poses of all keyframes.

To increase the number of valid loop closures and further improve the mapping accuracy, we extend our loop detection beyond image feature querying by also revisiting the inactive map.
Specifically, after performing tracking based on the active map $\mathcal{M}_A$ for each incoming frame, we additionally render images using the inactive Gaussian splats $\{ \boldsymbol{g} \in \mathcal{M}_I \}$. 
If the area of valid region in the rendered opacity image~$\boldsymbol{O}_i$ exceeds a threshold~$a_v$, indicating that the robot has observed part of the historical map. 
In this case, we count the occurrence numbers of all the closest observing frame ID $t_i^{c}$ among all observed Gaussian splats and accordingly select the keyframe with the highest count as the loop closure candidate.
We then input this candidate and the current frame into MASt3R, applying the same selection and tracking pipeline as described earlier.
\begin{figure}[t]
  \centering
  \includegraphics[width=1.0\linewidth]{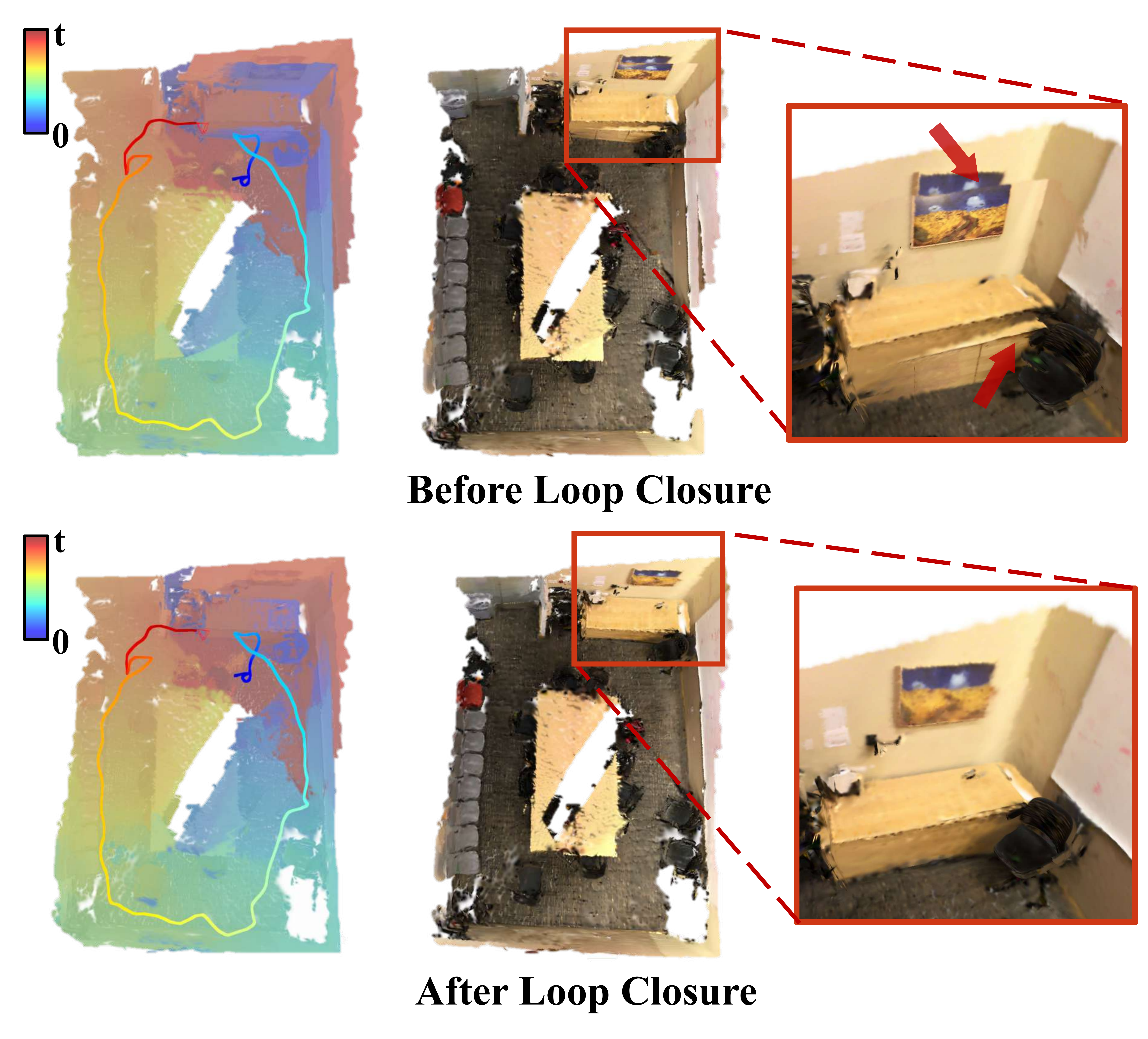}
  \caption{Result of loop closure and map correction. The left figures illustrate the associated keyframe ID of each Gaussian splat and the camera trajectory. 
  Both IDs and the trajectory are colored by time. The right figures compare the reconstruction results before and after loop closure. It can be observed that the map suffers from severe drift and misalignment before the loop correction (highlighted by red arrows). After that, the map structure becomes cleaner and more consistent.} 
  \label{fig:loop_comparison}
  \vspace{-10pt}
\end{figure}

\subsection{Map Correction}
\label{sec:correction}

After pose graph optimization, each keyframe $K_i$ with pose $\boldsymbol{T}_i$ is updated with an optimized pose increment:
\begin{equation}
  \Delta \boldsymbol{T}_i = \boldsymbol{T}_i^{o} \boldsymbol{T}_i^{-1},
\end{equation}
where $\boldsymbol{T}_i$ is the original pose, and $\boldsymbol{T}_i^{o}$ is the optimized pose.

For each Gaussian splat $\boldsymbol{g}_k \in \mathcal{M}$, let $f_c^k$ be the frame ID of its closest observed keyframe.
We apply the corresponding pose increment $\Delta \boldsymbol{T}_{f_c^k}$ to update the Gaussian's position $\boldsymbol{\mu}_k$ and orientation $\boldsymbol{R}_k$:
\begin{equation}
 \boldsymbol{\mu}_k' = \Delta \boldsymbol{T}_{f_c^k} \boldsymbol{\mu}_k, 
 \boldsymbol{R}_k' = \Delta \boldsymbol{R}_{f_c^k} \boldsymbol{R}_k,
\end{equation}
where $\boldsymbol{\mu}_k$ and $\boldsymbol{R}_k$ represent the original position and rotation of the Gaussian $\boldsymbol{g}_k$, and $\boldsymbol{x}_k'$ and $\boldsymbol{R}_k'$ are the updated values.
As illustrated in \figref{fig:loop_comparison}, our method ensures that the entire Gaussian map is deformed consistently with the optimized keyframe poses, preserving spatial coherence.

\section{Experimental Evaluation}
\label{sec:exp}

%
The main focus of this work is a rendering-based RGB-D SLAM system for building geometrically accurate and globally consistent radiance fields using 2D Gaussian splatting.

%
We present our experiments to show the capabilities of our method and analyze its performance. The results of our experiments also support our key claims, which are 
(i) Our proposed 2DGS-SLAM system demonstrates higher tracking accuracy than state-of-the-art rendering-based methods and traditional dense SLAM methods based on TSDF or surfel representations.
(ii) Our method outperforms 3DGS-based approaches in terms of surface reconstruction quality. The incorporation of the efficient loop closure mechanism ensures more globally consistent reconstruction results.
At the same time, our method also achieves comparable or superior image rendering quality, making the reconstruction result well-suited for downstream tasks.
(iii) Our 2DGS-SLAM is much more efficient in terms of runtime than other rendering-based SLAM systems with loop closures and generates a more compact map representation.
\subsection{Experimental Setup}

\begin{table}[t]
  \caption{Hyperparameters of our approach}
  \centering

  \begin{tabular}{c|c|c}
    \toprule
    symbol  & value  & description  \\
    \midrule
    \rowcolor{gray!15}
    \multicolumn{3}{l}{Tracking and Keyframe Selection, Sec.~\ref{sec:Tracking}} \\
    $c_k$  & 0.9 & covisibility threshold for keyframe selection \\
    $d_k$  & 15 $(cm)$ & distance threshold for keyframe selection \\
    $n_{\text{iter}}$ & 120 & maximum number of iterations \\
    \midrule
    \rowcolor{gray!15}
    \multicolumn{3}{l}{Mapping, Sec.~\ref{sec:Mapping}} \\
    $\lambda_{c}$ & $0.125$ & weight of the SSIM loss \\
    $w_{d}$ & $0.5$ & weight of depth loss \\
    $w_{n}$ & $0.02 $ & weight of normal loss \\
    $N_a$ & $3$ & active map training frames per iteration \\
    $N_i$ & $2$ & inactive map training frames per iteration \\
    \midrule
    \rowcolor{gray!15}
    \multicolumn{3}{l}{Loop Detection and Relocalization, Sec.~\ref{sec:Loop detection}} \\
    $s_r$  & 0.025 & similarity score threshold for image retrieval \\
    $c_s$  & 3.0 & mean confidence score threshold \\
    $a_v$  &0.5 & valid region threshold for revisiting loop \\
    $\delta_o$ & 0.5 & overlap ratio threshold  \\
    \bottomrule
  \end{tabular}

  \label{tab:parameters}
  \vspace{-10pt}
\end{table}

\subsubsection{Datasets}

We conduct our experiments on three public datasets that are widely adopted for performance evaluation in rendering-based SLAM methods as well as self-recorded data from a mobile robot. 
These datasets are the synthetic dataset Replica~\cite{straub2019arxiv-Replica}, and two real datasets, TUM-RGBD~\cite{sturm2012iros} and ScanNet~\cite{dai2017cvpr}.
The Replica dataset provides ground-truth camera poses along with an accurate mesh of the target environment. 
The TUM-RGBD dataset includes accurate camera poses captured using a motion capture system, while the reference poses in the ScanNet dataset are provided by BundleFusion~\cite{dai2017tog}.
It is worth noting that the depth images in the Replica dataset are rendered directly from the mesh and therefore free from noise. 
In contrast, both TUM-RGBD and ScanNet datasets are captured using consumer-grade structured-light-based RGB-D sensors, which introduce noticeable motion blur and depth measurement noise, presenting additional challenges for rendering-based SLAM algorithms. 
In addition to the three public datasets mentioned above, to evaluate the performance of our method on a real robotic platform, we recorded data using a wheeled robot in an indoor environment and conducted quantitative experiments on pose estimation.

\subsubsection{Implementation details}

We summarize the hyperparameters of our SLAM system, previously mentioned throughout the paper, in \tabref{tab:parameters}. 
These settings are kept consistent across all experiments. In addition, optimization-related parameters for 2D Gaussian splats, such as learning rates for different components, are also fixed for all datasets.
Due to variations in depth sensor accuracy, however, we adjust the tracking depth loss weight $\lambda_d$ and the tracking success threshold $\varepsilon_t$ individually for each dataset.
The pose graph optimization is carried out using GTSAM~\cite{dellaert2012factor}, employing the Levenberg-Marquardt method with a maximum iteration limit of 50.
We implement our system mainly using PyTorch, and all reported experiments are conducted on an NVIDIA A6000 GPU.

\begin{table*}[t]
  \caption{Absolute Trajectory Error (ATE) on the Replica dataset, reported in centimeters. \textbf{LC} denotes that loop closure is enabled. We highlight the best results in \textbf{bold} and the second best results are \underline{underscored}. }
  \centering
  \resizebox{0.9\textwidth}{!}{
    \begin{tabular}{c|c|c|cccccccc|c}
      \toprule
      \textbf{Method} & \textbf{Map Representation} & \textbf{LC} &\texttt{Rm 0} & \texttt{Rm 1} & \texttt{Rm 2} & \texttt{Off0} & \texttt{Off1} & \texttt{Off2} & \texttt{Off3} & \texttt{Off4} &  \textbf{Avg.} \\  
      \midrule
      NICE-SLAM~\cite{zhu2022cvpr} & feature grids &\ding{55} & 0.97 & 1.31 & 1.07 & 0.88 & 1.00 & 1.06 & 1.10 & 1.13 & 1.06 \\
      GO-SLAM~\cite{zhang2023iccv} & feature grids &\ding{51} & 0.34 & 0.29	& 0.29 & 0.32	& 0.30 & 0.39	& 0.39 & 0.46 & 0.35  \\
      E-SLAM~\cite{johari2023cvpr} & feature planes &\ding{55}  & 0.71 & 0.70 &	0.52 & 0.57 &	0.55 & 0.58 &	0.72 & 0.63 & 0.63 \\
      Point-SLAM~\cite{sandstrom2023iccv}  & feature points &\ding{55}  & 0.61 & 0.41	& 0.37 & 0.38 &	0.48 & 0.54 &	0.69 & 0.72 & 0.52 \\
      Loopy-SLAM~\cite{liso2024cvpr}  & feature points &\ding{51} & 0.24 & 0.24 & 0.28 & 0.26 & 0.40 & 0.29 & 0.22 & 0.35 & 0.29 \\
      PIN-SLAM~\cite{pan2024tro}     & feature points &\ding{55} & 0.27 & 0.31 & \underline{0.13} & \underline{0.22} & 0.30 & 0.28 & 0.16 & 0.28 & 0.24 \\
      RTG-SLAM~\cite{peng2024siggraph} & 3DGS &\ding{51} &\underline{0.20} &\underline{0.18} &\underline{0.13} &\underline{0.22} &\underline{0.12} &\underline{0.22} &0.20 &\underline{0.19} &\underline{0.18}\\
      MonoGS~\cite{matsuki2024cvpr-monogs}  & 3DGS &\ding{55} & 0.33 & 0.22 & 0.29 & 0.36 &	0.19 & 0.25 & \underline{0.12} & 0.81 & 0.32 \\
      SplaTAM~\cite{keetha2024cvpr-splatam}  & 3DGS &\ding{55} &  0.31 & 0.40 & 0.29 & 0.47 & 0.27 & 0.29 & 0.32 & 0.72 & 0.38\\
      Gaussian-SLAM~\cite{yugay2023arxiv}  & 3DGS &\ding{55} & 0.29 & 0.29 & 0.22 & 0.37 & 0.23 & 0.41 & 0.30 & 0.35 & 0.31 \\
      LoopSplat~\cite{zhu2025threedv-loopsplat}   & 3DGS &\ding{51}  & 0.28 & 0.22 &	0.17 & \underline{0.22} &	0.16 & 0.49 &	0.20 & 0.30 & 0.26 \\
      \midrule
      \bf{2DGS-SLAM (ours)} & 2DGS &\ding{51} & \bf{0.06} & \bf{0.08} & \bf{0.10}   & \bf{0.04} &	\bf{0.07} & \bf{0.07} &  \bf{0.06}  & \bf{0.09} & \bf{0.07} \\
      \bottomrule
    \end{tabular}
  }
  \vspace{-5pt}
  \label{tab:Replica_tracking}
\end{table*}

After completing the pose estimation of all frames, we directly merge the active map $\mathcal{M}_A$ and inactive map $\mathcal{M}_I$ to form a complete scene representation, 
which is then used to evaluate both reconstruction and rendering quality.
Following prior works~\cite{sandstrom2023iccv, matsuki2024cvpr-monogs, zhu2025threedv-loopsplat, liso2024cvpr}, we incorporate a map refinement stage to further enhance reconstruction results. 
Specifically, we perform an additional optimization of the map using all keyframes for 26,000 iterations.

\subsection{Tracking Performance}
\label{subsec:tracking_performance}

\begin{table}[t]
  \caption{Absolute Trajectory Error (ATE) on the TUM dataset, reported in centimeters. \textbf{LC} denotes that loop closure is enabled. We highlight the best results in \textbf{bold} and the second best results are \underline{underscored}. We separately compare the performance of rendering-based methods and classical approaches.}
  \centering
  \small
  \resizebox{\linewidth}{!}{
  \begin{tabular}{c|c|ccccc|c} 
    \toprule
    \textbf{Method} & \textbf{LC} & \texttt{desk} & \texttt{desk2} & \texttt{room} & \texttt{xyz} & \texttt{office} &  \textbf{Avg.} \\  
    \midrule
    \rowcolor{gray!15}
    \multicolumn{8}{l}{\textit{Rendering-based approach}} \\
    NICE-SLAM~\cite{zhu2022cvpr} & \ding{55} & 4.26 & 4.99 & 34.49 & 6.19 & 3.87 & 10.76 \\
    E-SLAM~\cite{johari2023cvpr}  & \ding{55} & 2.47 & 3.69 &	29.73 & \bf{1.11} &	2.42 & 7.89 \\
    Point-SLAM~\cite{sandstrom2023iccv} & \ding{55} & 4.34 & 4.54	& 30.92 & 1.31 &	3.48 & 8.92 \\
    Loopy-SLAM~\cite{liso2024cvpr} & \ding{51} & 3.79 & \underline{3.38} & 7.03 & 1.62 & 3.41 & 3.85 \\
    MonoGS~\cite{matsuki2024cvpr-monogs}  & \ding{55} & \bf{1.59} & 7.03 & 8.55 & 1.44 &	\bf{1.49} & 4.02 \\
    SplaTAM~\cite{keetha2024cvpr-splatam} & \ding{55} & 3.35 & 6.54 & 11.13 & 1.24 & 5.16 & 5.48 \\
    Gaussian-SLAM~\cite{yugay2023arxiv} & \ding{55} & 2.73 & 6.03 & 14.92 & 1.39 & 5.31 & 6.08 \\
    LoopSplat~\cite{zhu2025threedv-loopsplat} & \ding{51} & 2.08 & 3.54 & \underline{6.24} & 1.58 & 3.22 & \underline{3.33} \\
    \bf{2DGS-SLAM (ours)} & \ding{51} & \underline{1.84} & \bf{2.76} & \bf{5.98}   & \underline{1.16} &	\underline{1.97} & \bf{2.74} \\
    \midrule
    \rowcolor{gray!15}
    \multicolumn{8}{l}{\textit{Classical SLAM approach}} \\
    Kintinuous~\cite{whelan2012rssws} & \ding{51} & 3.7 & 7.1 & 7.5 & 2.9 & 3.0 & 4.84 \\
    ElasticFusion~\cite{whelan2015rss} & \ding{51} & \underline{2.0} & 4.8 & 6.8 & 1.1 & \underline{1.7} & 3.28 \\
    ORB-SLAM2~\cite{mur-artal2017tro} & \ding{51} & \bf{1.6} & \bf{2.2} & \bf{4.7} & \bf{0.4} & \bf{1.0} & \bf{2.0} \\
    RTAB-Map~\cite{labbe2019jfr}  & \ding{51} & 2.9 & \underline{4.4} & \underline{6.6} & \underline{0.5} & 2.1 & 3.3 \\
    
    \bottomrule
  \end{tabular}
  }
  \label{tab:tum_tracking}
\end{table}

\begin{table}[t]
  \caption{Absolute trajectory error (ATE) on the ScanNet dataset (cm). \textbf{LC} denotes that loop closure is enabled. We highlight the best results in \textbf{bold} and the second best results are \underline{underscored}.}
  \centering
  \resizebox{\linewidth}{!}{
  \begin{tabular}{c|c|cccccccc|c} 
    \toprule
    \textbf{Method} & \textbf{LC} &\texttt{00} & \texttt{59} & \texttt{106} & \texttt{169} & \texttt{181} & \texttt{207} & \texttt{54} & \texttt{233} &  \textbf{Avg.} \\  
    \midrule
    NICE-SLAM &\ding{55} & 12.0 & 14.0 & 7.9 & 10.9 & 13.4 & 6.2 & 20.9 & 9.0 & 13.0 \\
    GO-SLAM  &\ding{51} & \underline{5.4} & 7.5	& \bf{7.0} & 7.7	& \bf{6.8} & 6.9	& \underline{8.8} & 4.8 & \bf{6.9}  \\
    E-SLAM &\ding{55}  & 7.3 & 8.5 &	7.5 & \bf{6.5} &	9.0 & \bf{5.7} &	36.3 & \bf{4.3} & 10.6 \\
    Point-SLAM  &\ding{55}  & 10.2 & 7.8	& 8.7 & 22.0 &	14.8 & 9.5 &	28.0 & 6.1 & 14.3 \\
    Loopy-SLAM &\ding{51} & \bf{4.2} & 7.5 & 8.3 & \underline{7.5} & 10.6 & 7.9 & \bf{7.5} & 5.2 & 7.7 \\
    MonoGS &\ding{55} & 9.8 & 32.1 & 8.9 & 10.7 &	21.8 & 7.9 & 17.5 & 12.4 & 15.2 \\
    SplaTAM &\ding{55} &  12.8 & 10.1 & 17.7 & 12.1 & 11.1 & 7.5 & 56.8 & 4.8 & 16.6\\
    Gaussian-SLAM &\ding{55} & 21.2 & 12.8 & 13.5 & 16.3 & 21.0 & 14.3 & 37.1 & 11.1 & 18.4 \\
    LoopSplat   &\ding{51}  & 6.2 & \underline{7.1} &	\underline{7.4} & 10.6 &	\underline{8.5} & 6.6 &	16.0 & \underline{4.7} & 8.4 \\
    \midrule
    \bf{2DGS-SLAM} &\ding{51} & 6.6 & \bf{6.9} &  \underline{7.1}  & \bf{6.5} &	\underline{8.2} & \underline{6.0} &  11.0  & \underline{4.7} & \underline{7.1} \\
    \bottomrule
  \end{tabular}
  }
  \label{tab:scannet_tracking}
  \vspace{-5pt}
\end{table}

The first experiment evaluate how well our approach estimates the camera poses and compare it to existing baselinws. 
The results of theis experiment support our first claim that our 2DGS-SLAM system demonstrates higher tracking accuracy.
We evaluate tracking performance on all three datasets using the ATE RMSE~\cite{sturm2012iros} as the metric.
Among them, the Replica dataset is widely adopted for benchmarking rendering-based SLAM systems. 
On this synthetic dataset, we compare our 2DGS-SLAM method with several state-of-the-art approaches based on NeRF~\cite{mildenhall2020eccv}, 3D Gaussian splatting, and neural signed distance fields.
As shown in \tabref{tab:Replica_tracking}, our proposed 2DGS-SLAM outperforms all baselines, achieving sub-millimeter tracking accuracy. 
The trajectory error of our method is only half that of the second-best method, RTG-SLAM~\cite{peng2024siggraph}, which estimates camera poses by integrating multi-level ICP with ORB-SLAM2~\cite{mur-artal2017tro}, demonstrating the advantage and potential of rendering-based methods compared to traditional approaches.
This strong performance is largely attributed to the high-quality, noise-free depth images provided by Replica, which enable our 2DGS representation to fully exploit the advantages of consistent depth rendering.
These results also validate the effectiveness of our rendering-based camera pose optimization approach.

For the tracking results on the TUM-RGBD dataset, in addition to rendering-based methods, we also compare against classical RGB-D SLAM approaches such as Kintinuous~\cite{whelan2012rssws}, ElasticFusion~\cite{whelan2015rss}, ORB-SLAM2~\cite{mur-artal2017tro}, and RTAB-Map~\cite{labbe2019jfr}.
As reported in \tabref{tab:tum_tracking}, among rendering-based methods, our 2DGS-SLAM outperforms all baselines in terms of average accuracy.
In smaller-scale sequences such as \texttt{desk}, \texttt{xyz}, and \texttt{office}, our approach performs on par with state-of-the-art methods.
For larger scenes like \texttt{room} and sequences with more motion blur such as \texttt{desk2}, benefiting from the strength of our efficient loop closure mechanism, our method achieves the best performance.
Compared to classical methods, 2DGS-SLAM demonstrates superior performance over dense fusion approaches such as Kintinuous and ElasticFusion, but still falls slightly short of ORB-SLAM2.
Furthermore, the ScanNet dataset poses additional challenges, as all eight sequences are captured in room-scale or multi-room-scale indoor environments, where robust loop closure becomes critical. 
As shown in \tabref{tab:scannet_tracking}, SLAM methods without explicit loop closure mechanisms, 
such as Point-SLAM~\cite{sandstrom2023iccv}, MonoGS~\cite{matsuki2024cvpr-monogs}, and SplaTAM~\cite{keetha2024cvpr-splatam}, suffer from significantly higher pose estimation errors.
Our method ranks second in average trajectory accuracy across all eight sequences, demonstrating the strength of our loop closure strategy.
It is worth noting that the best-performing method, GO-SLAM~\cite{zhang2023iccv}, relies heavily on optical flow-based DROID-SLAM~\cite{teed2021neurips} for tracking and loop closure.
Additionally, the ground-truth trajectories in ScanNet are generated by BundleFusion~\cite{dai2017tog} rather than a high-precision motion capture system, and thus the results on this dataset should be considered as indicative rather than definitive.
\vspace{-5pt}

\begin{figure*}[t]
  \centering
  \begin{minipage}{0.245\linewidth}
      \includegraphics[width=\linewidth]{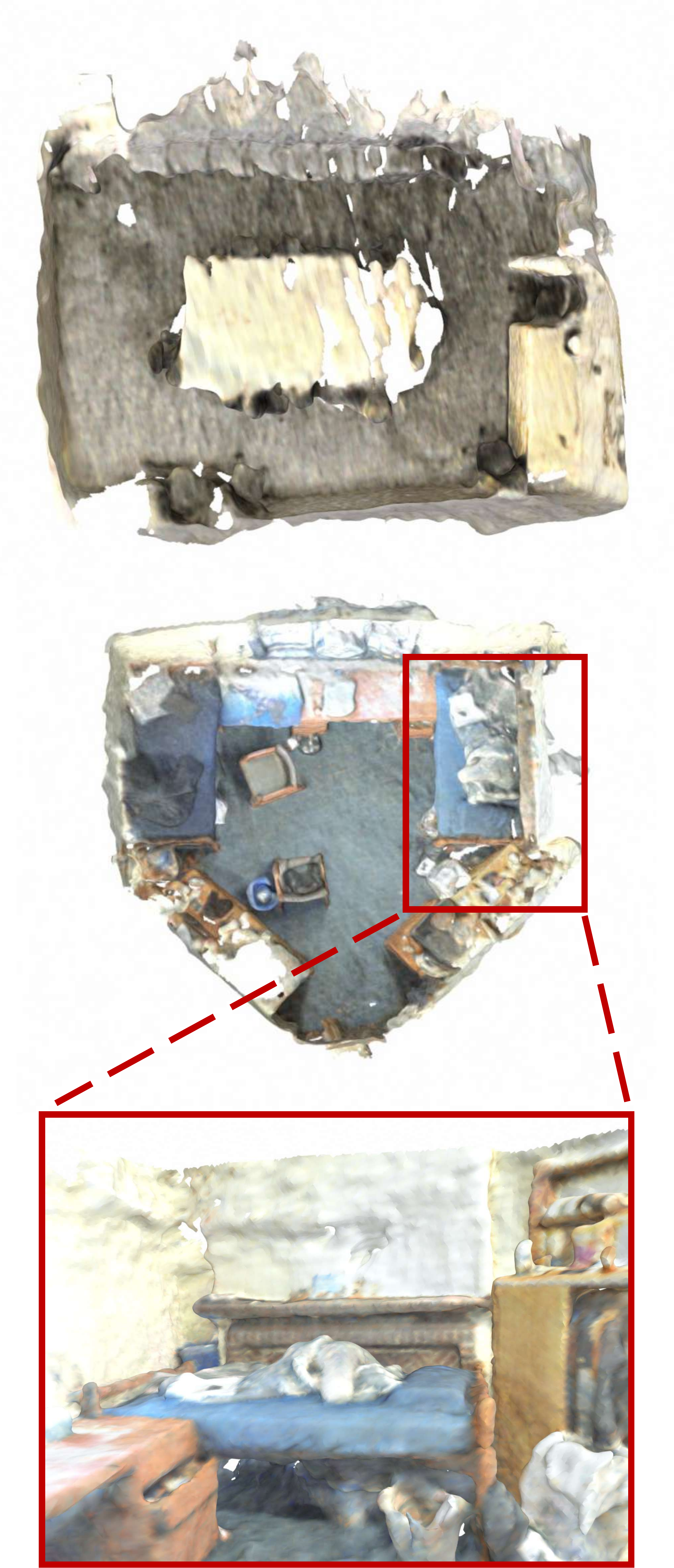}
      \centering
      {GO-SLAM~\cite{zhang2023iccv} } \\
  \end{minipage}
  \begin{minipage}{0.245\linewidth}
    \includegraphics[width=\linewidth]{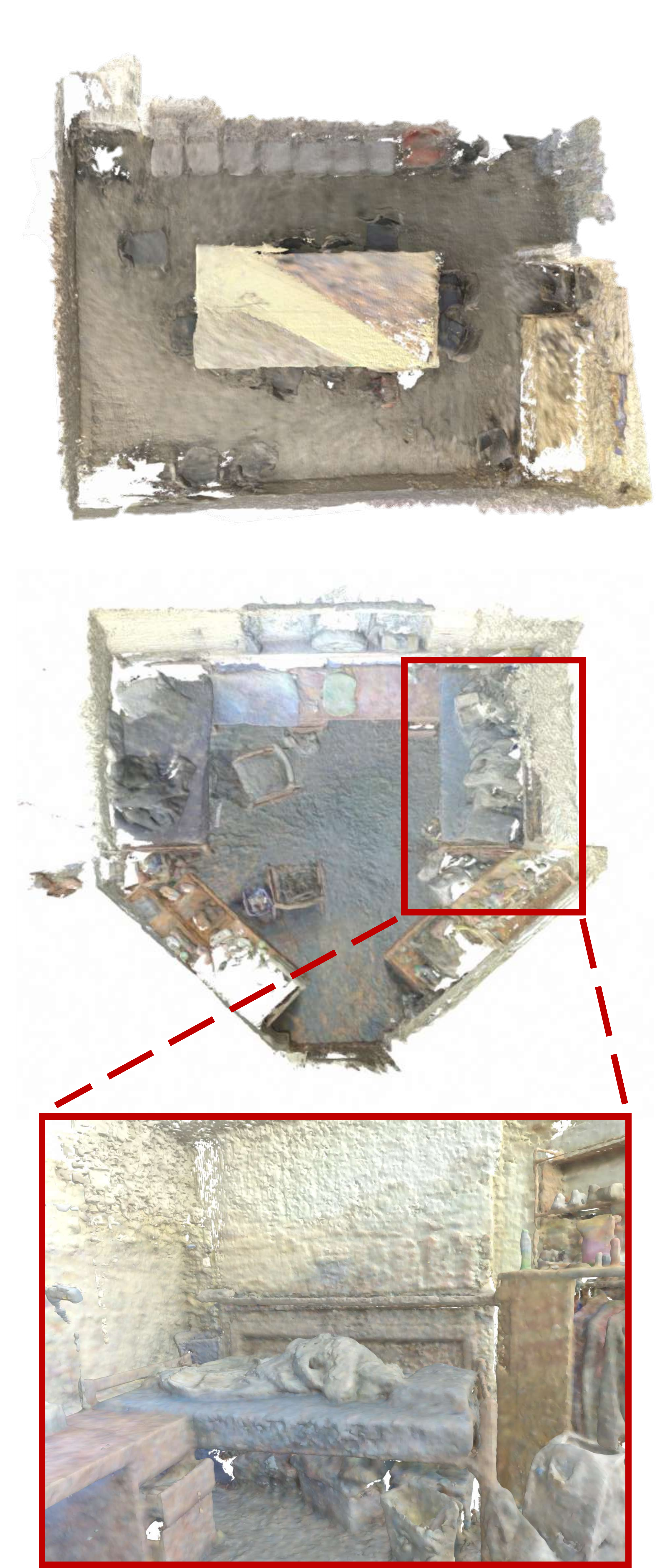}
    \centering
    { Loopy-SLAM~\cite{liso2024cvpr}} \\
  \end{minipage}
  \begin{minipage}{0.245\linewidth}
    \includegraphics[width=\linewidth]{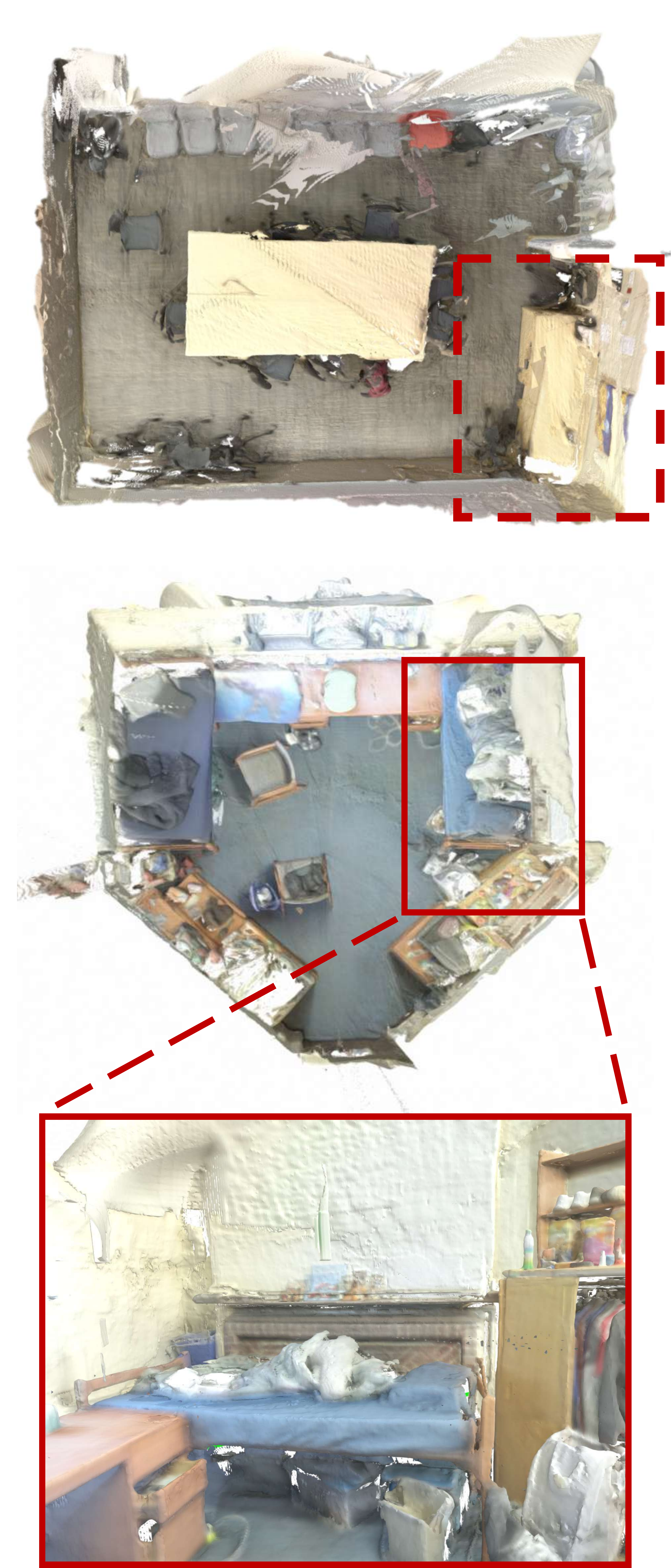}
    \centering
    { LoopSplat~\cite{zhu2025threedv-loopsplat}} \\
  \end{minipage}
  \begin{minipage}{0.245\linewidth}
      \includegraphics[width=\linewidth]{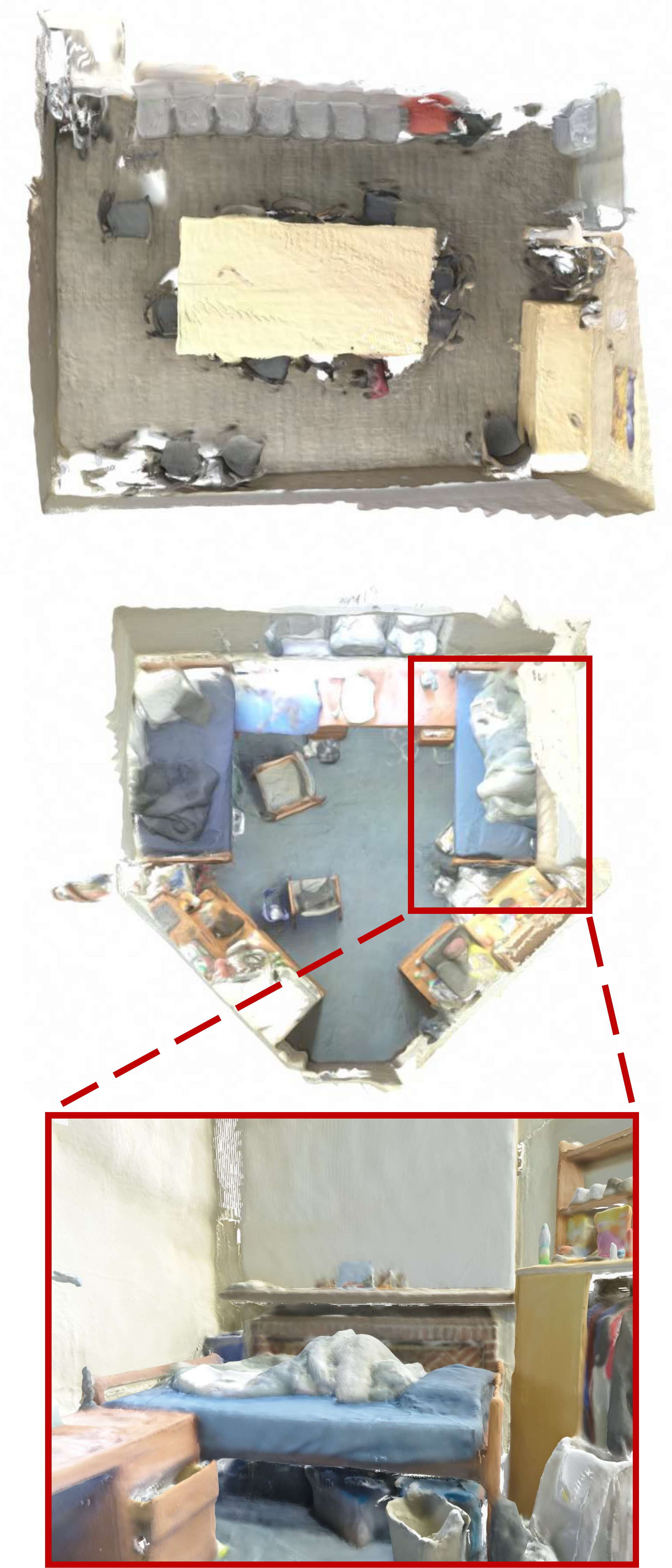}
      \centering
      {2DGS-SLAM (Ours)} \\
  \end{minipage}

  \caption{Qualitative comparison of reconstruction results on the ScanNet dataset. The first row presents the reconstructed meshes on sequence \texttt{0169}. We highlight the map duplication caused by LoopSplat's pose drift by a red dashed box.The second and third rows show the results on sequence \texttt{0233}, including both the overall meshes and zoomed-in local views. It can be observed that our 2DGS-SLAM achieves the most globally consistent and smooth reconstruction among all methods.}
  \label{fig:scannet_mesh}
  \vspace{-5pt}
\end{figure*}

\subsection{3D Reconstruction Performance}

\begin{table*}[t]
  \caption{Reconstruction comparison on the Replica dataset. We highlight the best results in \textbf{bold} and the second best results are \underline{underscored}. * indicates methods that use ground-truth depth for sampling.}
  \centering
  \resizebox{0.9\textwidth}{!}{
    \begin{tabular}{c|c|c|cccccccc|c}
    \toprule
    \textbf{Method} & \makecell{\textbf{Map}\\\textbf{Representation}} & \textbf{Metric} &\texttt{Rm 0} & \texttt{Rm 1} & \texttt{Rm 2} & \texttt{Off0} & \texttt{Off1} & \texttt{Off2} & \texttt{Off3} & \texttt{Off4} & \textbf{Avg.} \\
    \midrule
    \multirow{2}{*}{NICE-SLAM~\cite{zhu2022cvpr}} 
    & \multirow{2}{*}{feature grids} 
    & Depth L1[cm]$\downarrow$ & 1.81 & 1.44 & 2.04 & 1.39 &  1.76 & 8.33 & 4.99 & 2.01 & 2.97 \\
    & & F1 [\%]$\uparrow$ & 45.0 & 44.8 & 43.6 & 50.0 & 51.9 & 39.2 & 39.9 & 36.5 & 43.9 \\
    \midrule
    \multirow{2}{*}{E-SLAM~\cite{johari2023cvpr}}
    & \multirow{2}{*}{feature planes}
    & Depth L1[cm]$\downarrow$ & 0.97 & 1.07 & 1.28 & 0.86 &  1.26 & 1.71 & 1.43 & 1.06 & 1.18 \\
    & & F1 [\%]$\uparrow$ & 81.0 & 82.2 & 83.9 & 78.4 & 75.5 & 77.1 & 75.5 & 79.1 & 79.1 \\
    \midrule
    \multirow{2}{*}{Loopy-SLAM*~\cite{liso2024cvpr}}
    & \multirow{2}{*}{feature points}
    & Depth L1[cm]$\downarrow$ & \bf{0.30} & \bf{0.20} & \bf{0.42} & \bf{0.23} &  \underline{0.46} & \bf{0.60} & \bf{0.37} & \bf{0.24} & \bf{0.35} \\
    & & F1 [\%]$\uparrow$ & \bf{91.6} & \bf{92.4} & \underline{90.6} & \bf{93.9} & \bf{91.6} & \underline{88.5} & \bf{89.0} & \bf{88.7} & \bf{90.8} \\
    \midrule
    \multirow{2}{*}{SplaTAM~\cite{keetha2024cvpr-splatam}}
    & \multirow{2}{*}{3DGS}
    & Depth L1[cm]$\downarrow$ & 0.43 & 0.38 & 0.54 & 0.44 &  0.66 & 1.05 & 1.60 & 0.68 & 0.72 \\
    & & F1 [\%]$\uparrow$ & 89.3 & 88.2 & 88.0 & 91.7 & 90.0 & 85.1 & 77.1 & 80.1 & 86.1 \\
    \midrule
    \multirow{2}{*}{\makecell{Gaussian-\\SLAM~\cite{yugay2023arxiv}}}
    & \multirow{2}{*}{3DGS}
    & Depth L1[cm] $\downarrow$ & 0.61 & 0.25 & 0.54 & 0.50 & 0.52 & 0.98 & 1.63 & 0.42 & 0.68 \\
    & & F1 [\%] $\uparrow$      & 88.8 & 91.4 & 90.5 & 91.7 & 90.1 & 87.3 & 84.2 & 87.4 & 88.9 \\
    \midrule
    \multirow{2}{*}{LoopSplat ~\cite{zhu2025threedv-loopsplat}}
    & \multirow{2}{*}{3DGS}
    & Depth L1[cm]$\downarrow$ &  0.39 &  0.23 &  0.52 &  0.32 &  0.51 &   \underline{0.63} &   1.09 &  0.40 &   0.51\\
    & & F1 [\%]$\uparrow$ & 90.6 & \underline{91.9} & \bf{91.1} & \underline{93.3} & \underline{90.4} & \bf{88.9} & \underline{88.7} & \underline{88.3} & \underline{90.4} \\
    \midrule
    \multirow{2}{*}{\makecell{\textbf{2DGS-SLAM}\\\textbf{(ours)}}}
    & \multirow{2}{*}{2DGS}
    & Depth L1[cm] $\downarrow$  & \underline{0.34} & \underline{0.21} & \underline{0.43} & \underline{0.27} & \bf{0.41} & 1.08 & \underline{0.67} & \underline{0.28} & \underline{0.46} \\
    & & F1 [\%] $\uparrow$  & \underline{90.8} & 91.6 & \underline{90.6} & 93.1 & 90.1 & 87.0 & 87.6 & 87.5 & 89.7 \\
    \bottomrule
    \end{tabular}
  }
  \label{tab:Replica_reconstruction}
\end{table*}

\begin{table*}[t]
  \caption{Rendering performance comparison on the Replica dataset. We report three metrics: PSNR [dB], SSIM, and LPIPS. The best results are highlighted in \textbf{bold}, and the second best results are \underline{underscored}.}
  \centering
  \resizebox{0.8\textwidth}{!}{
    \begin{tabular}{l|c|cccccccc|c}
    \toprule
    \textbf{Method} & \textbf{Metric} &\texttt{Rm 0} & \texttt{Rm 1} & \texttt{Rm 2} & \texttt{Off0} & \texttt{Off1} & \texttt{Off2} & \texttt{Off3} & \texttt{Off4} & \textbf{Avg.} \\
    \midrule
    \multirow{3}{*}{Point-SLAM~\cite{sandstrom2023iccv}} 
    & PSNR $\uparrow$ & 32.40 & 34.08 & 35.50 &  38.26 &  39.16 &  33.99 &  33.48 & 33.49 & 35.17 \\
    & SSIM $\uparrow$      & \underline{0.974} & \underline{0.977} & 0.982 & \underline{0.983} & \underline{0.986} & 0.960 & 0.960 & \underline{0.979} & \underline{0.975} \\
    & LPIPS $\downarrow$   & 0.113 & 0.116 & 0.110 & 0.118 & 0.156 & 0.132 & 0.142 & 0.124 & 0.126 \\
    \midrule
    \multirow{3}{*}{SplaTAM~\cite{keetha2024cvpr-splatam}} 
    & PSNR $\uparrow$ & 32.86 & 33.89 & 35.25 & 38.26 &  39.17 & 31.97 & 29.70 & 31.81 & 34.11 \\
    & SSIM $\uparrow$ & \bf{0.980} & 0.970 & 0.980 & 0.980 & 0.980 & \underline{0.970} & 0.950 & 0.970 & 0.970 \\
    & LPIPS $\downarrow$ & 0.070 & 0.100 & 0.080 & 0.090 & 0.090 & 0.090 & 0.120 & 0.150 & 0.100 \\
    \midrule
    \multirow{3}{*}{MonoGS~\cite{matsuki2024cvpr-monogs}} 
    & PSNR $\uparrow$ & \underline{34.83} & \underline{36.43} & \underline{37.49} & 39.50 & \underline{42.09} & \underline{36.24} & \bf{36.70} & 36.07 & \underline{37.50} \\
    & SSIM $\uparrow$      & 0.954 & 0.959 & 0.965 & 0.971 & 0.977 & 0.964 & 0.963 & 0.957 & 0.960 \\
    & LPIPS $\downarrow$   & \underline{0.068} & \underline{0.076} & 0.075 & \underline{0.072} & \underline{0.055} & \underline{0.078} & \underline{0.065} & 0.099 & 0.070 \\
    \midrule
    \multirow{3}{*}{LoopSplat ~\cite{zhu2025threedv-loopsplat}} 
    & PSNR $\uparrow$ &  33.07 &  35.32 &  36.16 &  \underline{40.82} &   40.21 &   34.67 &   35.67 &    \underline{37.10} &   36.63\\
    & SSIM $\uparrow$ & 0.973 & \bf{0.978} & \bf{0.985} & \bf{0.992} & \bf{0.990} & \bf{0.985} & \bf{0.990} & \bf{0.989} & \bf{0.985} \\
    & LPIPS $\downarrow$ & 0.116 & 0.122 & 0.111 & 0.085 & 0.123 & 0.140 & 0.096 & 0.106 & 0.112 \\
    \midrule
    \multirow{3}{*}{\bf{2DGS-SLAM (ours)}}
    & PSNR $\uparrow$  & \bf{35.63} & \bf{37.09} & \bf{38.47} & \bf{43.14} & \bf{42.39} & \bf{36.33} & \underline{36.16} & \bf{38.8} & \bf{38.50} \\
    & SSIM $\uparrow$      & 0.965 & 0.968 & 0.973 & \underline{0.985} & 0.980 & 0.968 & \underline{0.966} & 0.971 & 0.972 \\
    & LPIPS $\downarrow$   & \bf{0.044} & \bf{0.048} & \bf{0.05} & \bf{0.029} & \bf{0.046} & \bf{0.049} & \bf{0.046} & \bf{0.049} & \bf{0.045} \\
    \bottomrule
    \end{tabular}
  }
  \label{tab:Replica_rendering}
\end{table*}

The second set of experiments evaluate the quality of the resulting model. The results support our second claim that our method outperforms 3DGS-based approaches in terms of surface reconstruction quality and global consistency.
We render depth images at keyframe poses using the global Gaussian splat map, followed by TSDF fusion~\cite{curless1996siggraph} to obtain the final reconstructed mesh.
We conduct quantitative evaluations on the Replica dataset, which provides ground-truth meshes for all sequences. 
Two commonly used metrics, Depth L1 error and F1 score are employed for the evaluation.
Depth L1 error measures the difference between the reconstructed and ground-truth meshes by rendering depth images from 1,000 randomly sampled camera poses and computing the per-pixel L1 distance. 
The F1 score~($F1$) evaluates the geometric accuracy of the mesh by jointly considering precision ($P$) and recall $(R)$, and is calculated as their harmonic mean: $F1 = 2\frac{PR}{P+R}$.
Here, precision ($P$) denotes the percentage of points on the predicted mesh that lie within 1\,cm of any point on the ground-truth mesh, 
while recall ($R$) measures the percentage of ground-truth points that are similarly close to the predicted mesh.
Our evaluation setup is consistent with previous works~\cite{zhu2022cvpr, sandstrom2023iccv, liso2024cvpr, zhu2025threedv-loopsplat}.
We select both Gaussian splatting-based and NeRF-style volume rendering-based methods as baselines for our quantitative experiments on the Replica dataset.

As shown in \tabref{tab:Replica_reconstruction}, in terms of Depth L1 error, our method ranks second, behind NeRF-based method Loopy-SLAM~\cite{liso2024cvpr}, outperforms other Gaussian Splatting-based methods. For the F1-score, our approach comes third, following Loopy-SLAM and LoopSplat~\cite{zhu2025threedv-loopsplat}. 
It is worth noting that, during depth rendering, \emph{Loopy-SLAM requires ground-truth depth to guide its sampling process.} While this contributes to its high accuracy on synthetic data, it limits the method’s applicability in real-world scenarios where depth measurements are noisy.
On the other hand, LoopSplat does not maintain a global map representation. Instead, it continuously generates local submaps during operation. For mesh reconstruction, LoopSplat renders depth images from these submaps, typically dozens per scene, and performs TSDF fusion. Since each submap only covers a limited range of viewpoints, the rendered depths tend to closely resemble the original input depth images. While this approach performs well on synthetic data, it often leads to excessive artifacts and map inconsistencies in real-world environments due to the lack of global information fusion. Moreover, maintaining a large number of overlapping submaps significantly increases memory consumption and complicates downstream robotic tasks such as planning.
\figref{fig:scannet_mesh} shows our qualitative results on the ScanNet dataset. We can observe that, in real-world environments with challenging lighting conditions and noisy depth measurements, Loopy-SLAM, which performs best on synthetic datasets, produces noticeably coarse mesh reconstructions. Similarly, LoopSplat suffers from issues such as more artifacts and map inconsistencies.
In contrast, our approach demonstrates superior global consistency and produces smoother surface reconstructions in real-world datasets.

\subsection{Rendering Quality}

The next set of experiments is designed to evaluate the rendering quality of our method. The results support the second part of our second claim, i.e., our approach enables high-fidelity rendering suitable for online robotic applications.
We evaluate the rendering quality of our method by computing the differences between the rendered images at all training views and their corresponding input images.
The evaluation metrics include peak signal-to-noise ratio (PSNR), structural similarity (SSIM)~\cite{wang2004tip}, and learned perceptual image patch similarity (LPIPS)~\cite{zhang2018cvpr-lpips}.
For baselines, we select the state-of-the-art NeRF-based methods, Point-SLAM~\cite{sandstrom2023iccv}, as well as Gaussian splatting based method, including SplaTAM~\cite{keetha2024cvpr-splatam}, MonoGS~\cite{matsuki2024cvpr-monogs}, and LoopSplat~\cite{zhu2025threedv-loopsplat}. 
We conduct quantitative evaluations on the Replica and ScanNet datasets. As shown in \tabref{tab:Replica_rendering}, 2DGS-SLAM achieves the best PSNR and LPIPS scores on the Replica dataset, with its SSIM score also being on par with other Gaussian splatting-based methods.
On the real-world ScanNet dataset, our method ranks second in average metric scores, with PSNR and SSIM worse than LoopSplat.
However, it is important to note that LoopSplat employs complex post-processing to merge its submaps.
Specifically, after completing pose estimation for all frames, LoopSplat first performs TSDF fusion using depth images rendered from different submaps to obtain the global mesh, then initializes a new set of Gaussian splats from the vertices of resulting mesh and optimizes them using all RGB-D keyframes for 30,000 iterations to generate a global radiance field. 
To isolate the impact of post-processing, we report extra rendering results of our method, MonoGS and LoopSplat without any map refinement on \tabref{tab:scannet_rendering2}. 
Since LoopSplat stores sub-maps instead of a unified global map, we directly merged its sub-maps using their respective poses to construct a global Gaussian splat map.
The results show that our method outperforms the baselines in terms of PSNR, SSIM, and LPIPS, and maintains competitive performance compared to the results obtained with map refinement.
Due to the severe pose drift, which can be seen in \tabref{tab:scannet_tracking}, MonoGS struggles to reconstruct a reliable radiance field for larger scenes online. 
Meanwhile, LoopSplat does not maintain a globally consistent map, leading to significant artifacts in the accumulated Gaussian splat submaps and making it unsuitable for high-quality rendering required by online robotic applications.

This observation is also supported by qualitative results on TUM dataset, as illustrated in the \figref{fig:tum_rendering}. Here, we compare the Gaussian splat maps obtained directly from each method after pose estimation, without any post-processing applied to the maps. As shown, LoopSplat's rendering results suffer from severe artifacts. Moreover, the normal renderings reveal that due to inconsistencies in 3DGS-based depth rendering, LoopSplat and MonoGS fail to produce smooth surface reconstructions. In comparison, our method not only achieves high-fidelity RGB renderings but also accurately reconstructs scene geometry. While SplaTAM achieves comparable reconstruction quality, it requires a much larger number of Gaussian splats than our approach. We provide a detailed comparison of memory and time consumption in the next section.

\begin{table}[t]
  \caption{Rendering performance comparison on the ScanNet dataset. We report three metrics: PSNR [dB], SSIM, and LPIPS. The best results are highlighted in \textbf{bold}, and the second best results are \underline{underscored}.}
  \centering
  \setlength{\tabcolsep}{3pt}
  \resizebox{0.48\textwidth}{!}{
    \begin{tabular}{l|l|cccccc|c}
      \toprule
      \textbf{Method} & \textbf{Metric} & \texttt{0000} & \texttt{0059} & \texttt{0106} & \texttt{0169} & \texttt{0181} & \texttt{0207} & \textbf{Avg.} \\
      \midrule
      \multirow{3}{*}{NICE-SLAM~\cite{zhu2022cvpr}} 
      & PSNR $\uparrow$ & 18.71 & 16.55 & 17.29 & 18.75 & 15.56 & 18.38  & 17.54 \\
      & SSIM $\uparrow$  & 0.641 & 0.605 & 0.646 & 0.629 & 0.562 & 0.646 & 0.621 \\
      & LPIPS $\downarrow$  & 0.561 & 0.534 & 0.510 & 0.534 & 0.602 & 0.552 & 0.548 \\
      \midrule
      \multirow{3}{*}{Point-SLAM~\cite{sandstrom2023iccv}} 
      & PSNR $\uparrow$  & 19.06 & 16.38 & 18.46 &  18.69 &  16.75 &  19.66  & 18.17\\
      & SSIM $\uparrow$  & 0.662 & 0.615 & 0.753 & 0.650 & 0.666 &0.696 & 0.673 \\
      & LPIPS $\downarrow$  & 0.515 & 0.528 & 0.439 & 0.513 & 0.532 & 0.500 & 0.504 \\
      \midrule
      \multirow{3}{*}{SplaTAM~\cite{keetha2024cvpr-splatam}} 
      & PSNR $\uparrow$  & 19.33 & \underline{19.27} & 17.73 & 21.97 &  16.76 & 19.8 & 19.14 \\
      & SSIM $\uparrow$  & 0.660 & \underline{0.792} & 0.690 & 0.776 & 0.683 & 0.696 & 0.716 \\
      & LPIPS $\downarrow$  & \bf{0.438} & \bf{0.289} & \underline{0.376} & \bf{0.281} & \bf{0.420} & \bf{0.341}  & \bf{0.358} \\
      \midrule
      \multirow{3}{*}{MonoGS~\cite{matsuki2024cvpr-monogs}} 
      & PSNR $\uparrow$  & 21.13 & 19.70 & 21.35 & 22.44& \underline{22.02} & 20.95 & 21.26 \\
      & SSIM $\uparrow$  & 0.723 & 0.722 & 0.808 & 0.781 & 0.814 & 0.725 & 0.762 \\
      & LPIPS $\downarrow$  & 0.448 & 0.436 & 0.339 & 0.362 & \underline{0.432} & 0.459 & \underline{0.412} \\
      \midrule
      \multirow{3}{*}{LoopSplat ~\cite{zhu2025threedv-loopsplat}} 
      & PSNR $\uparrow$  &  \bf{24.99} &  \bf{23.23} &  \bf{23.35} &  \bf{26.80} &  \bf{24.82} &  \bf{26.33} &   \bf{24.92}\\
      & SSIM $\uparrow$  & \bf{0.840} & \bf{0.831} & \bf{0.846} & \bf{0.877} & \bf{0.824} & \bf{0.854} & \bf{0.845} \\
      & LPIPS $\downarrow$  & 0.450 & \underline{0.400} & 0.409 & 0.346 & 0.514 & 0.430 & 0.425 \\
      \midrule
      \multirow{3}{*}{\bf{2DGS-SLAM}}
      & PSNR $\uparrow$  & \underline{23.36} & 19.00 & \underline{20.53} & \underline{24.67} & 21.27 & \underline{23.71} & \underline{22.09} \\
      & SSIM $\uparrow$  & \underline{0.767} & 0.729 & \underline{0.795} & \underline{0.796} & \underline{0.821} & \underline{0.779} & \underline{0.781} \\
      & LPIPS $\downarrow$  & \underline{0.440} & 0.444 & \bf{0.357} & \underline{0.362} & 0.485 & \underline{0.425} & 0.418 \\
      \bottomrule
      \end{tabular}
  }
  \label{tab:scannet_rendering}
\end{table}

\begin{table}[t]
  \caption{Rendering performance comparison on the ScanNet dataset. All the reported results are evaluated from the raw Gaussians splatting map without any refinement. We report three metrics: PSNR [dB], SSIM, and LPIPS. The best results are highlighted in \textbf{bold}, and the second best results are \underline{underscored}.}
  \centering
  \setlength{\tabcolsep}{3pt}
  \resizebox{0.48\textwidth}{!}{
    \begin{tabular}{l|l|cccccc|c}
      \toprule
      \textbf{Method} & \textbf{Metric} & \texttt{0000} & \texttt{0059} & \texttt{0106} & \texttt{0169} & \texttt{0181} & \texttt{0207} & \textbf{Avg.} \\
      \midrule
      \multirow{3}{*}{MonoGS~\cite{matsuki2024cvpr-monogs}} 
      & PSNR $\uparrow$  & \underline{15.40} & \underline{15.98} & \bf{18.34} & \underline{18.75} & \underline{15.43} & \underline{16.34} & \underline{16.70} \\
      & SSIM $\uparrow$  & \underline{0.597} & \underline{0.591} & \underline{0.701} & \underline{0.683} & \underline{0.642} & \underline{0.651} & \underline{0.644} \\
      & LPIPS $\downarrow$  & \underline{0.646} & \underline{0.591} & \underline{0.500} & \underline{0.525} & \underline{0.577} & \underline{0.577} & \underline{0.569} \\
      \midrule
      \multirow{3}{*}{LoopSplat ~\cite{zhu2025threedv-loopsplat}} 
      & PSNR $\uparrow$  & 12.35 & 12.95 & 10.26 &  10.86 & 11.47 &  13.17 &  11.84\\
      & SSIM $\uparrow$  & 0.413 & 0.411 & 0.318 & 0.495 & 0.541 & 0.504 & 0.447 \\
      & LPIPS $\downarrow$  & 0.840 & 0.724 & 0.798 & 0.791 & 0.698 & 0.704 & 0.759 \\
      \midrule
      \multirow{3}{*}{\bf{2DGS-SLAM}}
      & PSNR $\uparrow$  & \bf{21.95} & \bf{16.16} & \underline{17.71} & \bf{22.72} & \bf{19.74} & \bf{22.00} & \bf{20.05} \\
      & SSIM $\uparrow$  & \bf{0.740} & \bf{0.639} & \bf{0.710} & \bf{0.763} & \bf{0.793} & \bf{0.744} & \bf{0.731} \\
      & LPIPS $\downarrow$  & \bf{0.453} & \bf{0.501} & \bf{0.456} & \bf{0.392} & \bf{0.464} & \bf{0.435} & \bf{0.450} \\
      \bottomrule
      \end{tabular}
  }
  \label{tab:scannet_rendering2}
\end{table}


\begin{table}[t]
  \caption{Statistics of runtime and memory. We report three metrics: FPS, map size (MB), and peak GPU memory (MB). \textbf{LC} denotes that loop closure is enabled. The best results are highlighted in \textbf{bold}, and the second best results are \underline{underscored}.}
  \centering
  \setlength{\tabcolsep}{3pt}
  \resizebox{0.45\textwidth}{!}{
    \begin{tabular}{c|c|c|c|c}
      \toprule
      \textbf{Method} & \textbf{LC} & FPS (Hz) $\uparrow$ & Map size (MB) $\downarrow$ & GPU Memory (MB) $\downarrow$ \\
      \midrule
      Point-SLAM~\cite{sandstrom2023iccv}  &\ding{55} & 0.05  & 99.4 & \bf{8236}\\
      Loopy-SLAM~\cite{liso2024cvpr} &\ding{51} & 0.13  & 195.3 & 12475\\
      MonoGS~\cite{matsuki2024cvpr-monogs} &\ding{55} & \bf{1.92}  & \underline{13.2} & \underline{9062}\\
      SplaTAM~\cite{keetha2024cvpr-splatam} &\ding{55} & 0.18  & 213.1& 12939\\
      LoopSplat ~\cite{zhu2025threedv-loopsplat} &\ding{51} & 0.17  & 4608 & 9616\\
      \bf{2DGS-SLAM (ours)} &\ding{51} & \underline{0.92}  & \bf{9.7} & 10822\\
      \bottomrule
      \end{tabular}
  }
  \label{tab:runtime_memory}
  \vspace{-10pt}
\end{table}

\subsection{Runtime and Memory Evaluation}

The following experiment and results support the claim that our approach is more efficient in terms of runtime and produces a more compact map compared to the baselines. 
To compare the performance of different methods, we evaluate frames per second (FPS), calculated as the total number of frames in the sequence divided by the total time, as well as the memory usage of the map without post-processing and peak GPU memory consumption on the ScanNet sequence \texttt{scene0000}, which contains a total of 5,578 frames.
We selected main baselines from the previous experiments, including rendering-based methods such as Point-SLAM~\cite{sandstrom2023iccv}, Loopy-SLAM~\cite{liso2024cvpr}, MonoGS~\cite{matsuki2024cvpr-monogs}, SplaTAM~\cite{keetha2024cvpr-splatam}, and LoopSplat~\cite{zhu2025threedv-loopsplat}, for comparison.
As shown in~\tabref{tab:runtime_memory}, our method is only slower than MonoGS in terms of FPS. This is expected, as our approach involves additional tasks such as image feature extraction, loop closure detection, relocalization, and map updates, which are not exist in MonoGS as it does not incorporate loop closure. In comparison with other methods that do support loop closure, such as Loopy-SLAM~\cite{liso2024cvpr} and LoopSplat~\cite{zhu2025threedv-loopsplat}, our approach demonstrates significantly higher time efficiency, achieving a 6-7$\times$ speedup.

\begin{figure*}[t]
  \centering
  \begin{minipage}{0.24\linewidth}
      \includegraphics[width=\linewidth]{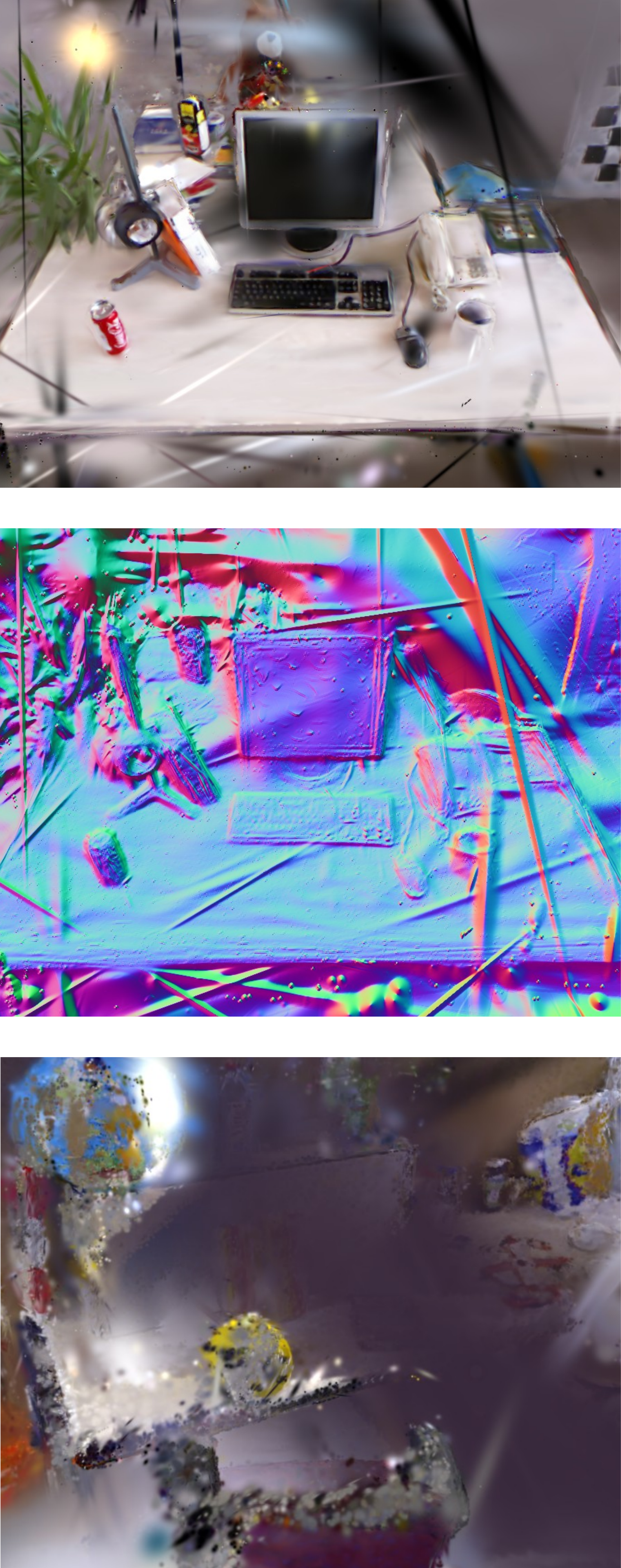}
      \centering
      {LoopSplat~\cite{zhu2025threedv-loopsplat} } \\
  \end{minipage}
  \begin{minipage}{0.24\linewidth}
    \includegraphics[width=\linewidth]{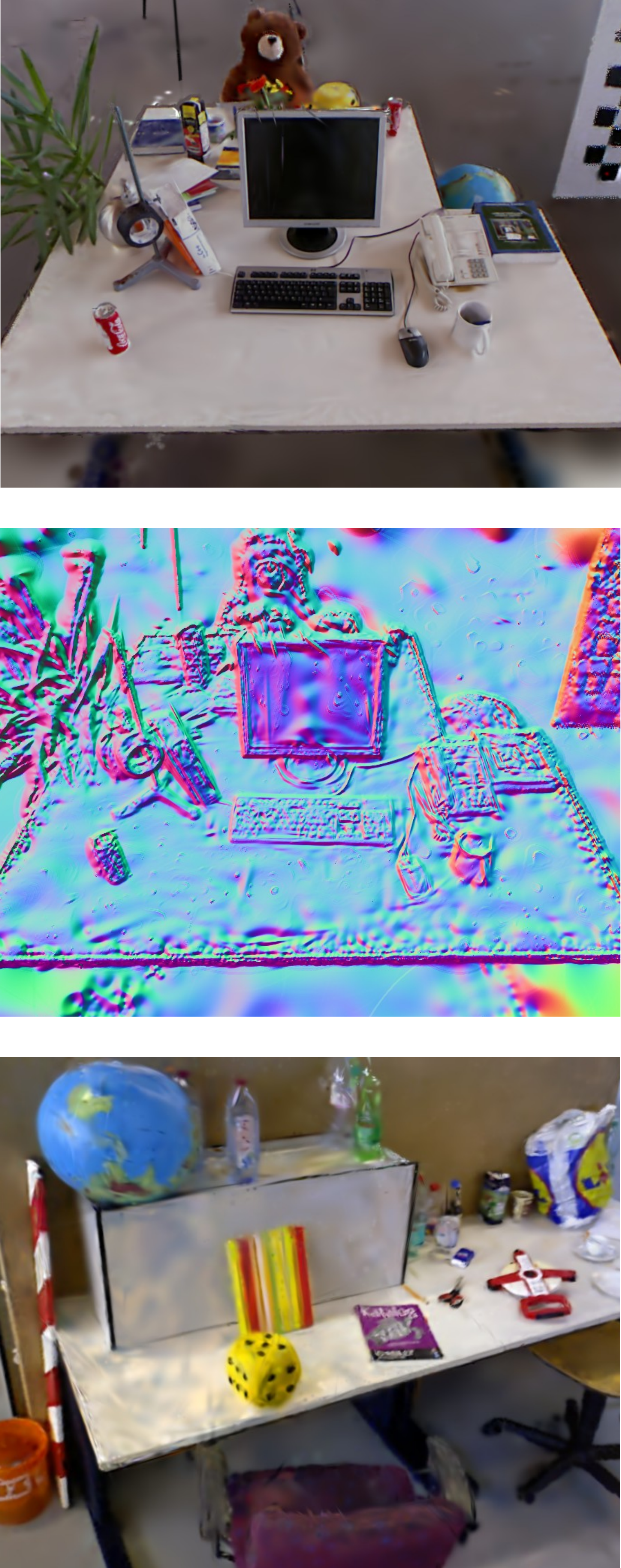}
    \centering
    { MonoGS~\cite{matsuki2024cvpr-monogs}} \\
  \end{minipage}
  \begin{minipage}{0.24\linewidth}
    \includegraphics[width=\linewidth]{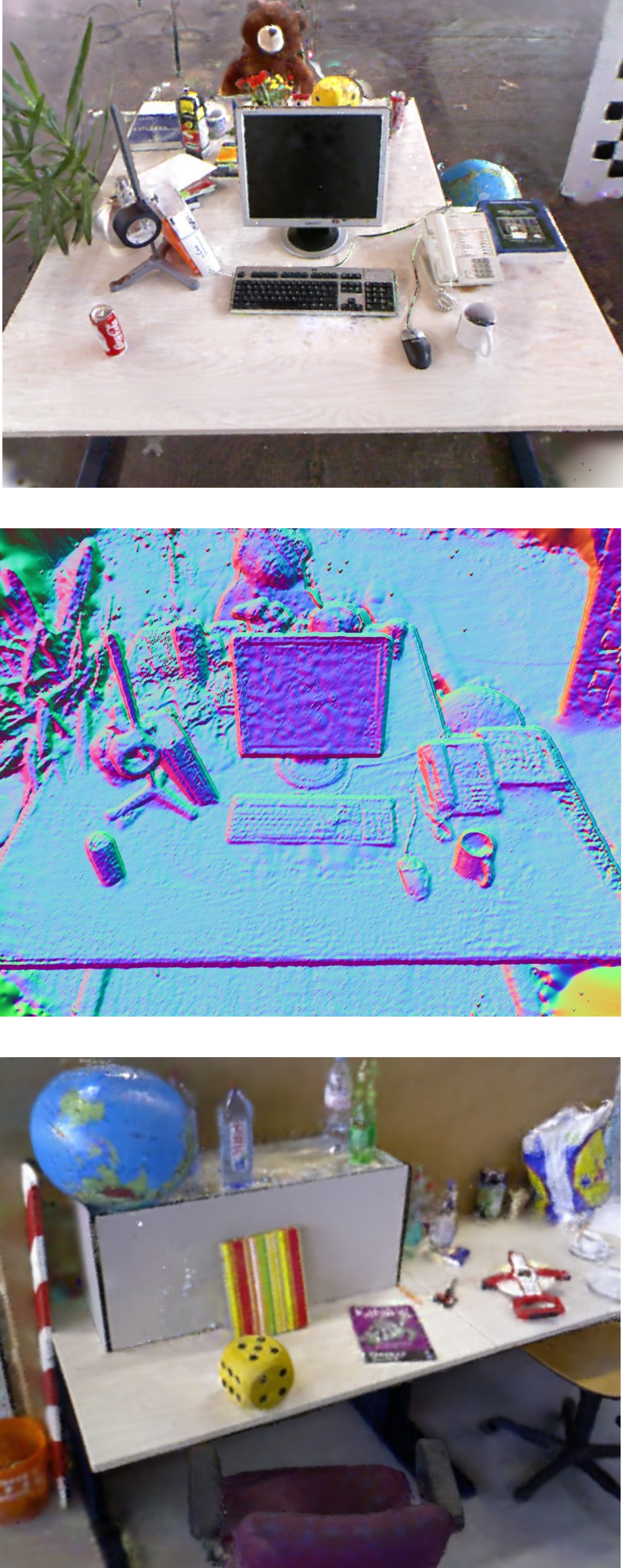}
    \centering
    { SplaTAM~\cite{keetha2024cvpr-splatam}} \\
  \end{minipage}
  \begin{minipage}{0.24\linewidth}
      \includegraphics[width=\linewidth]{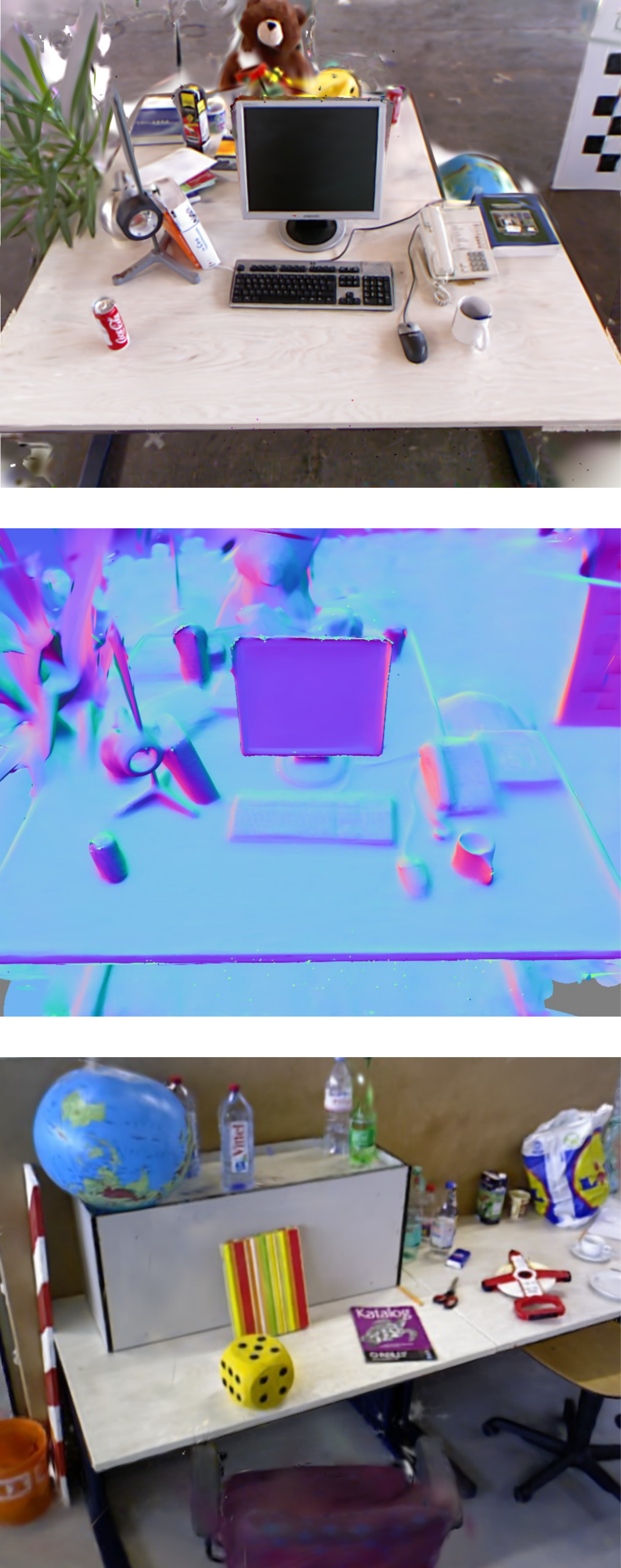}
      \centering
      {2DGS-SLAM (Ours)} \\
  \end{minipage}

  \caption{Qualitative comparison of Rendering results on the TUM dataset. For a fair comparison, we selected non-training views and used the raw Gaussian splatting maps from each method without any map refinement. The first and second rows show comparisons on sequence \texttt{fr2\_xyz}, including both RGB and normal renderings. As 3DGS does not support direct normal rendering, we compute normal images from the rendered depth images using \eqref{eq:normal_fromD} for visualization. The third row shows results on sequence \texttt{fr3\_office}. Our method achieves the most photorealistic RGB renderings and the smoothest normal image.}
  \label{fig:tum_rendering}
  \vspace{-6pt}
\end{figure*}

Additionally, and thanks to our efficient map management mechanism, our final map has the smallest memory footprint, suggesting that the number of redundant Gaussian splats in our system is much lower than in other Gaussian splatting-based methods. In contrast, due to the lack of removing redundant Gaussian splats, SplaTAM’s map memory usage is more than 20 times higher than ours. The continuously accumulating redundant splats also lead to a decrease in its pose estimation efficiency over time. Furthermore, since LoopSplat stores overlapping submaps rather than maintaining a global map, its memory usage for map storage is very high due to the accumulation of redundant splats.
In terms of peak GPU memory consumption, our method is slightly higher than LoopSplat. However, this is because LoopSplat offloads all submaps to disk in order to minimize runtime memory usage. Unfortunately, the frequent disk I/O and CPU-GPU data transfers significantly slow down its speed compared to ours.
In summary, when compared to other rendering-based SLAM methods with loop closure support, 2DGS-SLAM outperforms the baselines in both memory usage and runtime efficiency.

\subsection{Experiments on Self-recorded Robot Data}

To evaluate the effectiveness of our method in real-world robotic applications beyond publicly available datasets, we also collected data using a wheeled mobile robot equipped with Intel RealSense D455 RGB-D cameras in indoor environments. 
The experimental scenes include (1) \texttt{corridor}, a 20-meter-long straight corridor used to evaluate the robustness of our pose estimation in low-texture, repetitive environments; (2,3) \texttt{kitchen} and \texttt{office}, two rooms measuring approximately 7\,m $\times$ 6\,m, where the robot performs challenging maneuvers such as rapid pure rotations during recording. As shown in \figref{subfig:ipb_a}, we use AprilTags mounted on the ceiling to compute near ground-truth poses with approximately 1\,cm global accuracy for evaluation. It is also worth noting that, compared to the structured-light-based RGB-D cameras used in datasets such as ScanNet~\cite{dai2017tog} and TUM-RGBD~\cite{sturm2012iros}, the stereo-vision-based RealSense D455 typically produces noisier depth images.

As illustrated in \figref{subfig:ipb_b}, our method achieves high-quality scene reconstruction, demonstrating not only a high-fidelity radiance field but also smooth surface normal rendering. We further conducted quantitative pose estimation experiments, comparing our method with the main baselines evaluated in the aforementioned public datasets. As shown in \tabref{tab:ipb_tracking}, our method yields substantially lower average trajectory error than all baseline methods, highlighting its robustness to depth noise and rapid camera motion.
Moreover, our method successfully performs loop closures on both the \texttt{kitchen} and \texttt{office} sequences, significantly reducing pose drift compared to competing methods. Consistent with the observations in experiment~\ref{subsec:tracking_performance}, methods that lack loop closure support, such as Point-SLAM~\cite{sandstrom2023iccv}, MonoGS~\cite{matsuki2024cvpr-monogs}, and SplaTAM~\cite{keetha2024cvpr-splatam}, suffer from severe pose drift, making them unsuitable for room-scale reconstruction and real-world mobile robot applications. compared with rendering-based methods with loop closure capability, including Loopy-SLAM~\cite{liso2024cvpr} and LoopSplat~\cite{zhu2025threedv-loopsplat}, our approach demonstrates superior robustness in both motion estimation and loop closure, highlighting the practical value of our method in robotic applications.

\begin{figure}[t]
  \centering
  \begin{subfigure}[b]{0.46\textwidth}
    \centering
    \includegraphics[width=\textwidth]{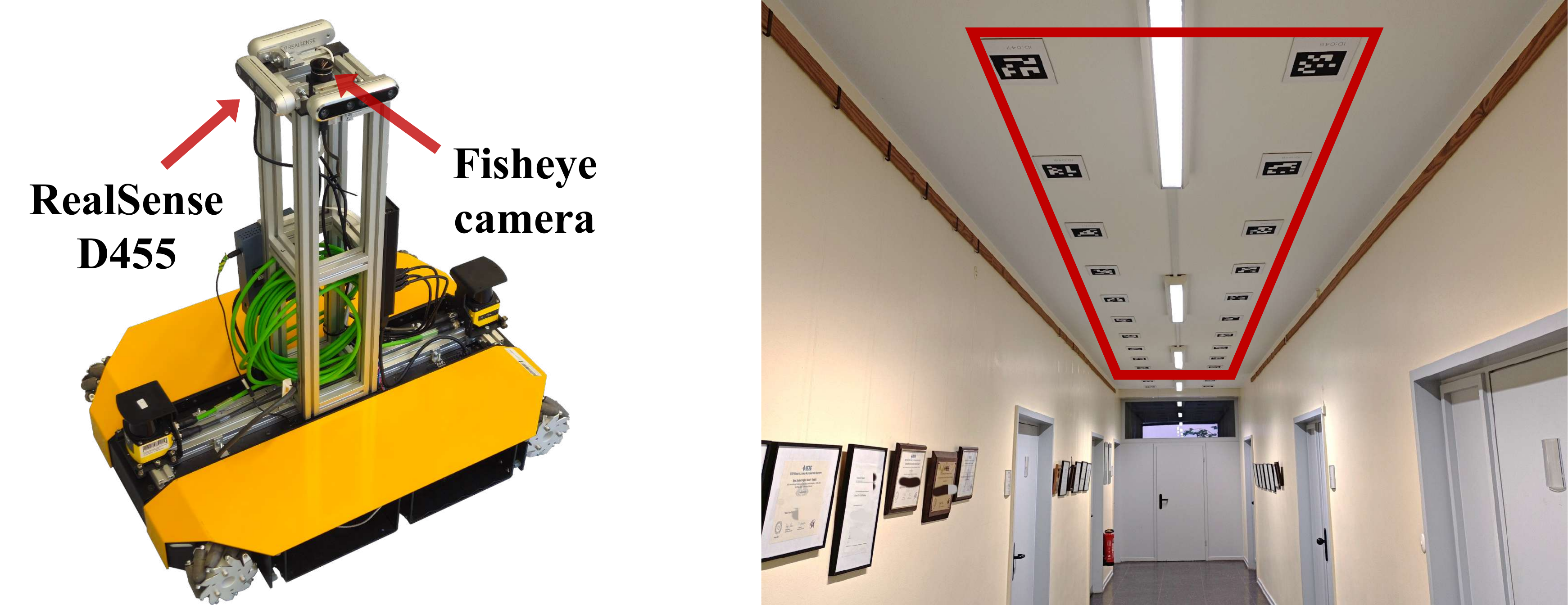}
    \caption{}\vspace{5pt}
    \label{subfig:ipb_a}
  \end{subfigure}
  \vfill
  \begin{subfigure}[b]{0.50\textwidth}
    \centering
    \includegraphics[width=0.9\textwidth]{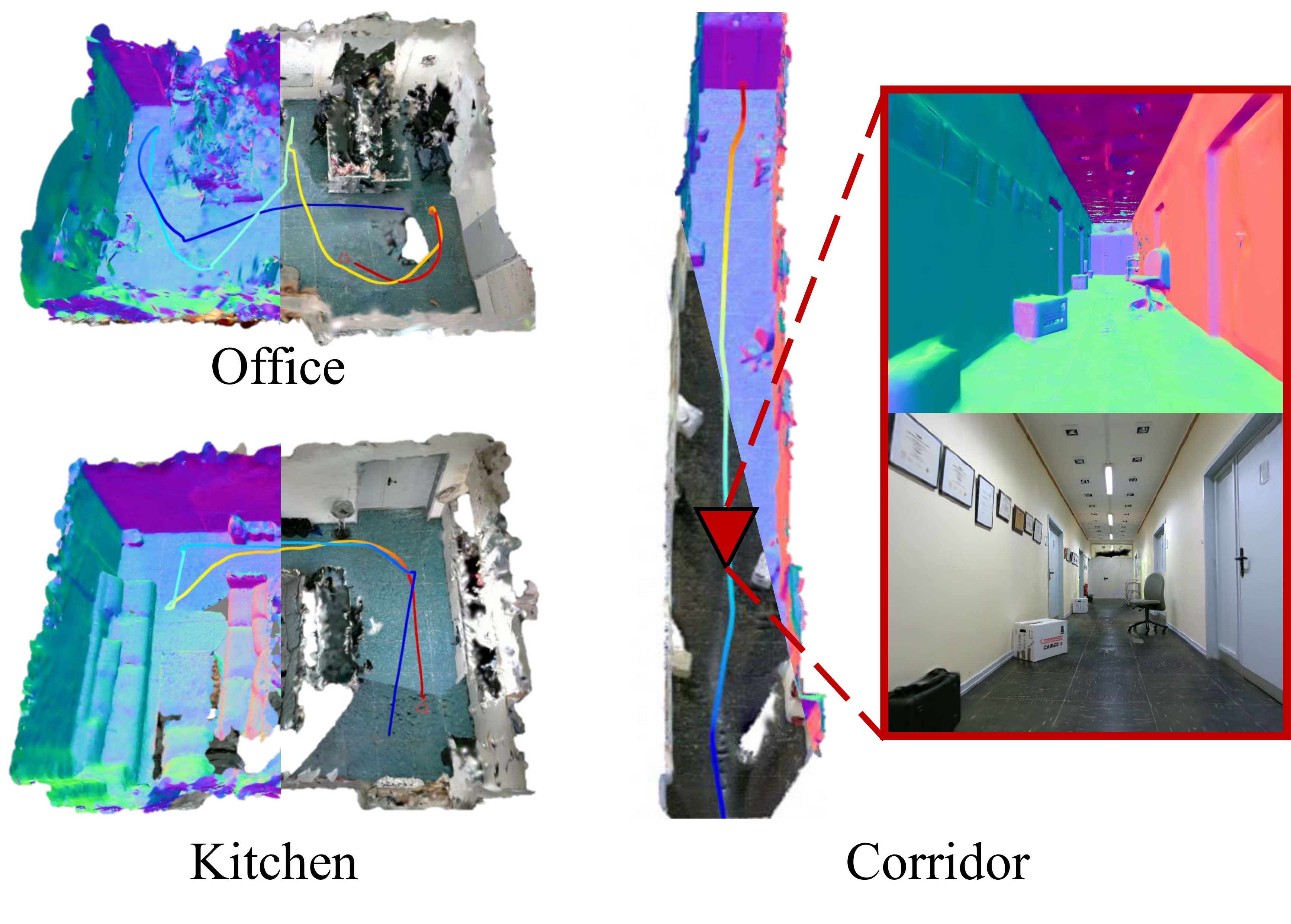}
    \vspace{-10pt}
    \caption{}
    \label{subfig:ipb_b}
  \end{subfigure}

  \caption{(a) The wheeled robot platform used in our experiments and AprilTags mounted on the ceiling. We use the fisheye camera installed on the robot to detect AprilTags for pose evaluation. (b)~Reconstructed Gaussian splat maps and camera trajectories of our method on three experimental scenes: \texttt{office}, \texttt{kitchen}, and \texttt{corridor}. For the \texttt{corridor}, we demonstrate zoomed-in views of both RGB and normal renderings. No map refinement was applied after tracking.}
  \label{fig:ipb}
  \vspace{-10pt}
\end{figure}

\begin{table}[t]
  \caption{Absolute trajectory error (ATE) on the self-recorded dataset (cm). \textbf{LC} denotes that loop closure is enabled. The best results are highlighted in \textbf{bold}, and the second best results are \underline{underscored}.}
  \centering
  \resizebox{\linewidth}{!}{
  \begin{tabular}{c|c|ccc|c} 
    \toprule
    \textbf{Method} & \textbf{LC} &\texttt{corridor} & \texttt{kitchen} & \texttt{office} & \textbf{Avg.} \\  
    \midrule
    Point-SLAM~\cite{sandstrom2023iccv}  &\ding{55}  & 30.8 & 15.9 & 23.9 & 23.5 \\
    Loopy-SLAM~\cite{liso2024cvpr} &\ding{51} & Failed & 63.9 & 7.9 & - \\
    MonoGS~\cite{matsuki2024cvpr-monogs} &\ding{55} & 9.5 & 17.6 & \underline{13.9} & 13.6 \\
    SplaTAM~\cite{keetha2024cvpr-splatam} &\ding{55} & 29.8 & 130.5 & 16.7 & 59.0\\ 
    LoopSplat~\cite{zhu2025threedv-loopsplat}   &\ding{51}  & \bf{2.1} & \underline{10.1} & 22.6 & 11.6 \\
    \midrule
    \bf{2DGS-SLAM (ours)} &\ding{51} & \underline{3.4} & \bf{6.8} & \bf{4.7} & \bf{5.0} \\
    \bottomrule
  \end{tabular}
  }
  \label{tab:ipb_tracking}
  \vspace{-10pt}
\end{table}

\section{Conclusion}
\label{sec:conclusion}

In this paper, we proposed 2DGS-SLAM, a novel RGB-D SLAM framework that enables globally consistent radiance field reconstruction based on 2D Gaussian splatting. Taking advantage of the consistent depth rendering of 2D Gaussian splatting, we propose an accurate camera tracking framework. We further introduced an efficient map management strategy and integrated a strong 3D foundation model MASt3R to enable robust loop closure detection and relocalization.
We implemented and evaluated our approach on different datasets and provided comparisons to other existing techniques and supported all claims made in this paper. The results demonstrate that our method achieves superior pose estimation accuracy compared to other rendering-based approaches, while delivering comparable or even better surface reconstruction quality. Moreover, our 2DGS-SLAM consistently outperforms 3D Gaussian splatting-based systems in terms of surface smoothness and global consistency. At the same time, our method maintains competitive image rendering quality with significantly improved efficiency compared with other rendering based method with loop closure support.


\section{Acknowledgements}
\label{sec:acknowledgements}
We thank Haofei Kuang and Niklas Trekel for providing the real-world robot-collected datasets along with ground-truth poses, and Liyuan Zhu for sharing the baseline results.

\bibliographystyle{plain_abbrv}

\bibliography{glorified,new}

\IfFileExists{./certificate/certificate.tex}{
\subfile{./certificate/certificate.tex}
}{}
\end{document}